\documentclass[10pt,journal,compsoc]{IEEEtran}

\usepackage[caption=false]{subfig}
\usepackage{overpic}
\usepackage{graphicx}
\usepackage{amsmath}
\usepackage{amssymb}
\usepackage{booktabs}
\usepackage{verbatim}
\usepackage{url}
\usepackage{multirow}
\usepackage[dvipsnames, table, xcdraw]{xcolor}
\usepackage{pifont}

\usepackage{hyperref}
\hypersetup{colorlinks=true, linkcolor=blue, anchorcolor=blue, citecolor=blue}

\newcommand{\cmark}{\ding{51}}%
\newcommand{\xmark}{\ding{55}}%

\usepackage{silence}
\definecolor{mygray}{gray}{.95}
\definecolor{myRed}{RGB}{219, 68, 55}
\definecolor{myGreen}{RGB}{15, 157, 88}
\definecolor{myBlue}{RGB}{66, 133, 244}

\def\ie{\emph{i.e.,~}}

\def\etc{\emph{etc}}

\newcommand{\figref}[1]{Fig.~\ref{#1}}
\newcommand{\tabref}[1]{Tab.~\ref{#1}}

\newcommand{\secref}[1]{Sec.~\ref{#1}}
\newcommand{\myPara}[1]{\vspace{5pt}\noindent\textbf{#1}}

\graphicspath{{./pami_figs/}}

\usepackage{makecell}

\newcommand{\addFig}[1]{}
\newcommand{\addFigs}[1]{}

\def\nameofbenchmark{Ref-COD}
\def\nameofdataset{R2C7K}
\def\camsubset{Camo-}
\def\refsubset{Ref-}
\def\nameofmethod{R2CNet}
\def\mainbranch{segmentation branch}
\def\refbranch{reference branch}
\def\genmoudule{Referring Mask Generation}
\def\enrmoudule{Referring Feature Enrichment}
\def\sOTa{state-of-the-art~}
\def\etc{\emph{etc}}

\def\salsuffix{Ref}

\def\nameofmodifycolor{black}   
\newcommand{\modifying}[1]{\textcolor{\nameofmodifycolor}{#1}}

\usepackage[nocompress]{cite}

\begin{document}

\title{Referring Camouflaged Object Detection}

\author{
        Xuying Zhang,
        Bowen Yin,
        Zheng Lin,
        Qibin Hou,~\IEEEmembership{Member,~IEEE},\\
        Deng-Ping Fan,~\IEEEmembership{Senior Member,~IEEE},
        Ming-Ming Cheng,~\IEEEmembership{Senior Member,~IEEE}

        \IEEEcompsocitemizethanks{
            \IEEEcompsocthanksitem X. Zhang, B. Yin, Q. Hou, D. Fan and M. Cheng are with VCIP, School of Computer Science, Nankai University, Tianjin, China (zhangxuying1004@gmail.com).   
            \IEEEcompsocthanksitem Z. Lin is with BNRist, Department of Computer Science and Technology, Tsinghua University, Beijing, China.       
            \IEEEcompsocthanksitem First two authors contributed equally.
            \IEEEcompsocthanksitem Z. Lin and Q. Hou are the corresponding authors.\\ (frazer.linzheng@gmail.com, andrewhoux@gmail.com).
            \IEEEcompsocthanksitem This research was supported by NSFC (NO. 62225604, No. 62276145), the Fundamental Research Funds for the Central Universities (Nankai University, 070-63223049). Computations were supported by the Supercomputing Center of Nankai University (NKSC).
        }
}

\IEEEtitleabstractindextext{%
\begin{abstract}
We consider the problem of referring camouflaged object detection (\nameofbenchmark{}), a new task that aims to segment specified camouflaged objects based on a small set of referring images with salient target objects.
We first assemble a large-scale dataset, called \nameofdataset{}, which consists of 7K images covering 64 object categories in real-world scenarios.
Then, we develop a simple but strong dual-branch framework, dubbed \nameofmethod{}, with a \refbranch{} embedding the common representations of target objects from referring images and a \mainbranch{} identifying and segmenting camouflaged objects under the guidance of the common representations. 
In particular, we design a \genmoudule{} module to generate pixel-level prior mask and a \enrmoudule{} module to enhance the capability of identifying specified camouflaged objects.
Extensive experiments show the superiority of our \nameofbenchmark{} methods over their COD counterparts in segmenting specified camouflaged objects and identifying the main body of target objects. 
Our code and dataset are publicly available at \url{https://github.com/zhangxuying1004/RefCOD}.
\end{abstract}

}

\maketitle
\IEEEdisplaynontitleabstractindextext

\IEEEpeerreviewmaketitle

\IEEEraisesectionheading{\section{Introduction}\label{sec:introduction}}

\IEEEPARstart{C}{amouflaged} object detection (COD), which aims to segment objects that are visually hidden in their surroundings, has been attracting more and more attentions~\cite{fan2022concealed,pang2022zoom, ji2023deep, sun2022boundary}. 
This research topic plays an important role in a wide range of real-world applications, \emph{e.g.}, medical image segmentation~\cite{fan2020pranet,fan2020inf}, surface defect detection~\cite{tabernik2020segmentation,le2020learning}, and pest detection~\cite{turkouglu2019plant}.
It is noteworthy that although there may be multiple camouflaged objects in a real scene,  we just want to find the specified ones in many applications.
A typical example should be the explorers are looking for some special species but most of them may be hidden deep together with other similar objects.
In this case, if we have some references about the targets, the finding process will become orientable and thus get easier.
As a result, it is promising to explore the COD research with references, which is abbreviated as \nameofbenchmark{} in this paper.

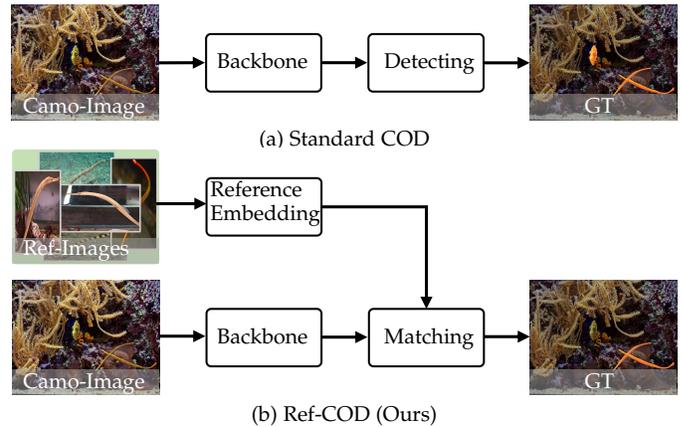
\begin{figure}[t]
  \centering
  \footnotesize
  \begin{overpic}[width=\linewidth]{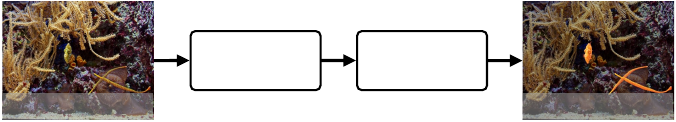}
    \put(2,1.3){\color{white}{Camo-Image}}
    \put(86,1.3){\color{white}{GT}}
    \put(31,8){Backbone}
    \put(56,8){Detecting}
  \end{overpic} 
  \\ (a) Standard COD \\
  \begin{overpic}[width=\linewidth]{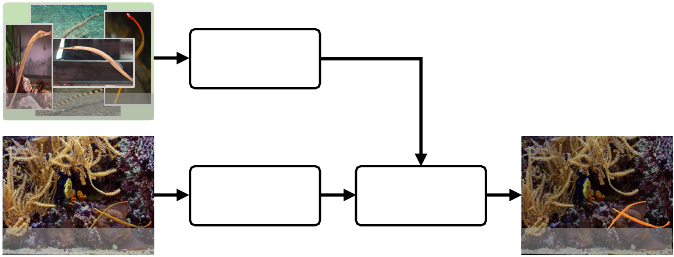}
    \put (2, 21.3){\color{white}{Ref-Images}}
    \put (2, 1.3){\color{white}{Camo-Image}}
    \put (86, 1.3){\color{white}{GT}}
    \put (30,30){Reference}
    \put (30,26.5){Embedding}
    \put (31,8){Backbone}
    \put (56,8){Matching}
  \end{overpic} 
  \\ (b) Ref-COD (Ours) \\
  \caption{Visual comparison between the standard COD and our 
    \nameofbenchmark{}. 
    Given an image containing multiple camouflaged objects, 
    the COD model tends to find all possible camouflaged objects 
    that are blended into the background without discrimination, 
    while the \nameofbenchmark{} model attempts to identify the specified 
    camouflaged objects under the condition of a set of referring images.
  }\label{fig:intro} 
\end{figure}

\nameofbenchmark{} leverages the referring information to guide the identification of specified camouflaged objects, which is consistent with that
of human visual perception to camouflaged objects~\cite{troscianko2009camouflage}.
And the key issue lies on which form of information is appropriate to be used as a reference.
Recent works have explored several forms of reference for image segmentation, \emph{e.g.}, referring expression segmentation with text reference~\cite{hu2016segmentation}, and few-shot segmentation with image reference~\cite{zhang2019canet}.
However, whether the annotated images containing specified camouflaged objects or detailed textual descriptions of camouflaged objects for existing images, the acquisition process is time-consuming and laborious, which hinders the transfer of these methods to COD. 
Considering that images with salient objects are readily available on the Internet, a straightforward question arises:
Is it possible for us to take advantage of such images to help better identify specified camouflaged objects?

Motivated by this question, we propose a novel \nameofbenchmark{} benchmark.
Our intention is to leverage the increasingly advanced research on salient object detection (SOD) to acquire common representations of target objects from referring images, which are used to guide the segmentation of specified camouflaged objects.
\figref{fig:intro} illustrates the task relationship between the standard COD and our \nameofbenchmark{}. 
In particular, our \nameofbenchmark{} transforms the process of COD from detecting aimlessly the differential objects in camouflage scenes to matching target objects with specified purposes.
To enable a comprehensive study on this new benchmark, we build a large-scale dataset, named \emph{\nameofdataset{}}, which contains a large number of samples without copyright disputes in real-world scenarios. 
The basic information of this dataset is as follows: 1) It has 7K images covering 64 object categories; 2) It consists of two subsets, \emph{i.e.}, the \camsubset subset composed of images containing camouflaged objects and the \refsubset subset composed of images containing salient objects; 3) The number of images for each category in the \refsubset subset is fixed, while the one in the \camsubset subset is not. 

To investigate the role of the referring information in \nameofbenchmark{}, we design a dual-branch network architecture and develop a simple but effective framework, named \emph{\nameofmethod{}}.
This framework includes a \refbranch{} and a \mainbranch{}.
The \refbranch{} aims to capture common representations of target objects from the referring images composed of salient objects, which will be used to identify the specified camouflaged objects. 
Particularly, we build a \genmoudule{} (RMG) module to generate pixel-level referring information. 
In this module, a dense comparison is performed between the common representations from the \refbranch{} and each position of the visual features from the \mainbranch{} to generate a referring prior mask.
However, there may exist variances in appearance between the camouflaged objects and the salient objects even though they belong to the same category, which may increase the difficulty of identifying accurate camouflaged objects.
To overcome this shortcoming, a dual-source information fusion strategy is employed to eliminate the information differences between two information sources. 
In addition, we also design a \enrmoudule{} (RFE) module to achieve the interaction among multi-scale visual features under the guidance of referring mask, and further highlight the target objects.

Extensive experiments are conducted to validate the effectiveness of our \nameofbenchmark{}. 
To be specific, we choose the \mainbranch{} with multi-scale feature fusion based on feature pyramid network (FPN)~\cite{lin2017feature} as the baseline COD model, and compare it with our \nameofmethod{} in terms of the common metrics~\cite{perazzi2012saliency,fan2017structure,fan2018enhanced,margolin2014evaluate} of COD research on the \nameofdataset{} dataset.
Remarkably, our \nameofmethod{} outperforms the baseline model by a large margin.  
Furthermore, we also apply the design of \nameofbenchmark{} on recent 7 \sOTa{} COD methods, and the \nameofbenchmark{} methods consistently surpass their COD counterparts without bells and whistles.
Besides, the visualization results also show quality predictions of the \nameofbenchmark{} methods (e.g., \nameofmethod{}) in the segmentation of specified objects and the identification of the main body of camouflaged objects over the COD  model (e.g., baseline). 

To sum up, the contributions of this paper can be summarized as follows:
\begin{itemize}
\item We propose a new benchmark, termed \nameofbenchmark{}. To the best of our knowledge, it is the first attempt to bridge SOD and COD and segment the specified camouflaged objects with salient object images.

\item We build a large-scale dataset, named \nameofdataset{}, which could help provide data basis and deeper insights for the \nameofbenchmark{} research.

\item We design a new framework for \nameofbenchmark{} research, dubbed \nameofmethod{}, whose excellent experimental results suggest that it could offer an effective solution to this novel topic. 

\end{itemize}

\section{Related Work}\label{sec:relatedwork}
In this section, we first present the research progress and existing problems of COD tasks. 
Then, we introduce the development history of the SOD topic. 
Finally, we describe referring object segmentation research with different forms of reference information.

\subsection{Camouflaged Object Detection}
Identifying camouflaged objects from a confusing scene is challenging due to the high similarity to their surroundings, diversity in scale, fuzziness in appearance, \emph{etc}. 
To solve this task, an increasing number of works have been proposed. 
One of the early systematic studies is presented by~\cite{fan2020camouflaged}, which publishes a large-scale dataset with high-quality annotation, namely COD10K, and a well-designed search and identify framework for this research. 
Later on, a large number of strategies, \emph{e.g.}, multi-scale feature fusion~\cite{sun2021context,chou2022finding,zhuge2022cubenet,pang2022zoom,chen2022camouflaged}, multi-stage refinement~\cite{Jia_2022_CVPR,fan2022concealed,zhang2022preynet,wang2021d}, graph learning~\cite{zhai2021mutual}, weakly supervised~\cite{he2022weakly}, uncertainty~\cite{li2021uncertainty,liu2022modeling,kajiura2021improving}, 
foreground and background separation~\cite{mei2021camouflaged,yin2022camoformer,zhai2022deep}
, and attention mechanism~\cite{zhang2022camouflaged,cheng2022attention} have been proposed to improve the performance of the COD models. 
More related works are presented in recent survey paper~\cite{fan2023advances}.
More recently, numerous works introduce additional information, \emph{e.g.}, boundary~\cite{zhu2022can,sun2022boundary,zhou2022feature,qin2021boundary,ji2022fast}, texture~\cite{zhu2021inferring,ren2021deep,ji2023deep}, frequency~\cite{Zhong_2022_CVPR,lin2023frequency}, and 
depth~\cite{zhang2021depth,xiang2021exploring,wu2022source,wu2022source}, as guidance to further improve the accuracy of the segmentation results. 
However, obtaining these forms of information of high quality from the existing camouflaged images requires expensive time and resource costs,
and they cannot explicitly tell the COD model what to segment.

In this paper, we propose a novel benchmark, which extracts common representations of target objects from readily available referring images as explicit semantic guidance to locate and segment the specified camouflaged objects.
This make our work quite different from the standard COD task.
%

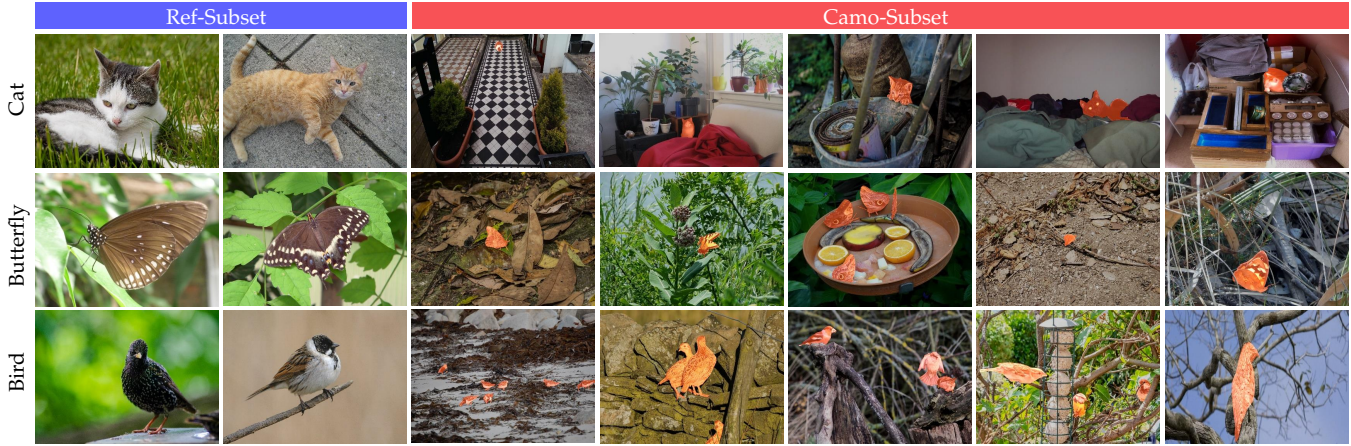
\begin{figure*}[t]
  \centering
  \footnotesize
  \begin{overpic}[width=0.96\linewidth]{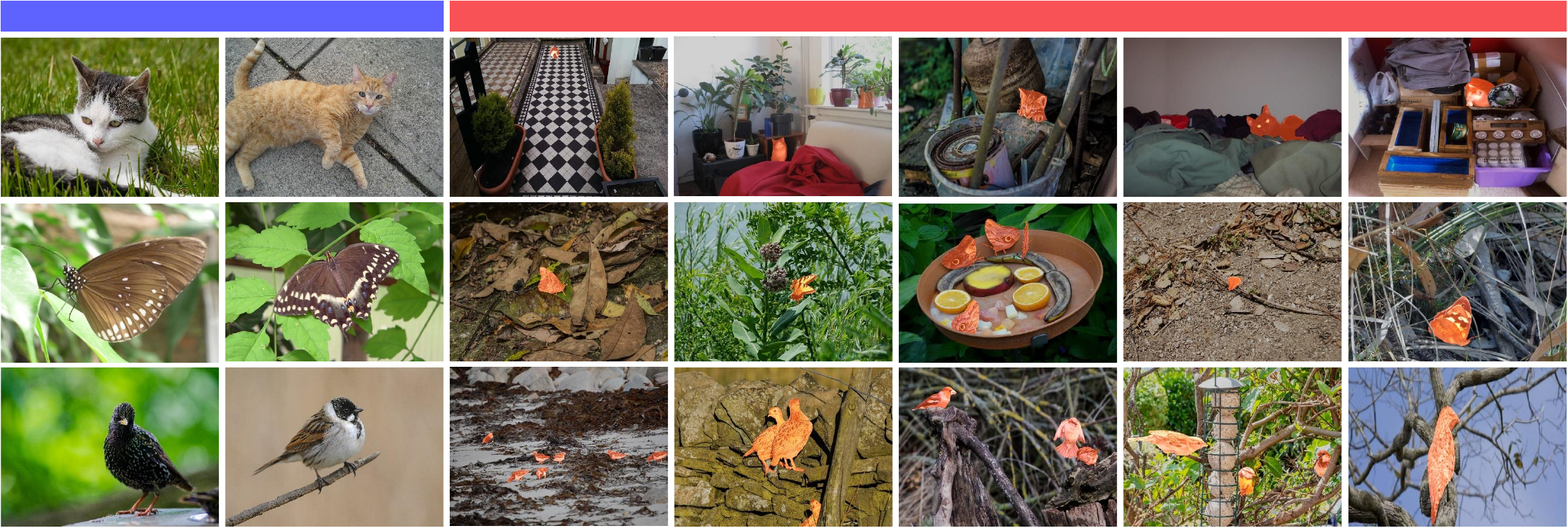}
    \put (10,32){{\color{white}{\refsubset Subset}}}
    \put (60,32){{\color{white}{\camsubset Subset}}}
    \put (-2,25){\rotatebox{90}{Cat}}
    \put (-2,12){\rotatebox{90}{Butterfly}}
    \put (-2,4){\rotatebox{90}{Bird}}
  \end{overpic}
  \caption{
    Examples from our \nameofdataset{} dataset. 
    Note that the camouflaged objects in \camsubset subset are masked 
    with their annotations in orange.
  }\label{fig:s2c7k-sample}
\end{figure*}

\subsection{Salient Object Detection}
Salient object detection (SOD) aims to capture the most attention-grabbing objects in the given image.
The development of this topic has witnessed two stages: 1) the traditional method stage and 2) the deep learning method stage.
During the former period, hand-crafted features~\cite{cheng2014global,wang2015deep,kim2016shape,he2015supercnn} from image patches, object proposals, and super-pixels, play the most important role.
Although these methods work well in some scenes, the extraction of these features is usually time-consuming. More importantly, their performance may sharply degrade in complex conditions.
With the prosperity of fully convolutional networks~\cite{long2015fully} and Transformers~\cite{vaswani2017attention,zhang2021rstnet,wu2022difnet,zhuge2022salient,ma2023towards}, SOD research enters the latter stage.
In this stage, U-shape structure~\cite{lin2017feature,ronneberger2015u}, multi-stage supervision~\cite{fan2021group,zhang2020gradient,hou2017deeply}, and attention mechanisms~\cite{liu2018picanet,chen2018reverse} are widely used in SOD methods to achieve more accurate pixel-wise predictions.

It is worth mentioning that SOD has extensive applications in various research communities, \emph{e.g.}, visual tracking~\cite{borji2012adaptive}, unsupervised segmentation~\cite{li2022exploring}, image compression~\cite{guo2009novel}, and content-aware image editing~\cite{cheng2010repfinder}, mainly because it can help find the objects or regions that can efficiently represent a scene. 
This property of SOD also inspires the reference idea of this paper.


\subsection{Referring Object Segmentation}
Referring Object Segmentation means segmenting visual objects from a given image under a certain form of references, \emph{e.g.}, image and text. 

Few-shot segmentation (FSS) explores object segmentation guided by annotated images containing objects of the same category. 
The models are trained on a large number of images whose pixels are labeled with base classes (query set) and perform dense pixel prediction on unseen classes given a few annotated samples (support set). 
Particularly, most existing FSS networks include two branches, \emph{i.e.}, support branch and query branch, to extract the features of support images and query images and achieve the interaction between them. 
The pioneering work of FSS research is proposed by~\cite{shaban2017one}, where the support branch directly predicts the weights of the last layer in the query branch for segmentation. 
Then, the masked average pooling operation is proposed by~\cite{zhang2020sg} to extract representative support features, which is widely adopted by subsequent works.
More recently, a large number of works~\cite{zhang2019canet,tian2020prior,lang2022learning} build powerful modules on the frozen backbone network to improve the adaptability of the models to unseen categories.

Referring Expression Segmentation (RES) explores object segmentation guided by a textual expression. 
RES aims to segment visual objects based on the expression, and the two-branch architecture is also adopted by the networks in this research. The first work is introduced by~\cite{hu2016segmentation}, where the visual and linguistic features are first extracted by a visual encoder and a language encoder respectively, and their concatenation is employed to generate the segmentation mask.
Subsequently, a series of methods based on multi-level visual features~\cite{li2018referring}, multi-modal LSTM~\cite{liu2017recurrent}, attention mechanism~\cite{shi2018key,zhou2021real}, collaborative network~\cite{luo2020multi} are incorporated in RES methods successively to generate more accurate results. 
In addition, the text descriptions are also adopted by~\cite{sun2020exploring} as references for the richness of image content to achieve a better fixation prediction. 


In this paper, the proposed \nameofbenchmark{} also belongs to a referring object segmentation task. 
However, different from the existing methods, the collection of its referring information does not take much effort. 
To be specific, it neither needs to collect rare and hard-to-label images containing target camouflaged objects  nor annotates detailed text descriptions for existing COD datasets, which is convenient for academia and industry to follow.

\section{Proposed Dataset}
The emergence of a series of datasets build the basis for carrying out artificial intelligence research, especially in the current data-hunger deep learning era. Besides, the quality of a dataset plays an important role in its lifespan as a benchmark, as stated in~\cite{perazzi2016benchmark,wang2018revisiting}. With this in mind, we build a large-scale dataset, named \nameofdataset{}, for the proposed \nameofbenchmark{} task. In this section, we introduce the construction process and statistics of this dataset, respectively. 

\subsection{Data Collection and Annotations}
To construct the \nameofdataset{} dataset, the first step is to determine which camouflaged objects to detect. 
To this end, we investigate the most popular datasets in COD research, \emph{i.e.}, COD10K~\cite{fan2022concealed}, CAMO~\cite{le2019anabranch}, and NC4K~\cite{lv2021simultaneously}.
Considering that COD10K is the largest and most comprehensively annotated camouflage dataset, we build the \camsubset subset of \nameofdataset{} mainly based on it.
Specifically, we eliminate a few unusual categories, \emph{e.g.} pagurian, crocodile-fish, \emph{etc}, and attain 4,966 camouflaged images covering 64 categories. 
For the images containing only one camouflaged object, we directly adopt the annotations provided by COD10K, and for other images containing multiple camouflaged objects, we erase the annotated pixels except for objects of the referring category.
Note that we also supplement 49 samples from NC4K for some categories due to their extremely small sample numbers.

Next, we construct the \refsubset subset of \nameofdataset{} according to the selected 64 categories. We use these category names as keywords and search 25 images that come from real-world scenarios and contain the desired salient objects from the Internet for each category. 
In particular, these referring images, which have no copyright disputes, are collected from Flickr and Unsplash.
For the details on the image collection scheme, we recommend the readers to refer to~\cite{zhou2017places}. 

Finally, we present the image and annotation samples of the \nameofdataset{} dataset in \figref{fig:s2c7k-sample}.

\begin{figure*}[t]
    \centering
    \footnotesize
    \subfloat[Object Area]{
        \includegraphics[width=.235\linewidth]{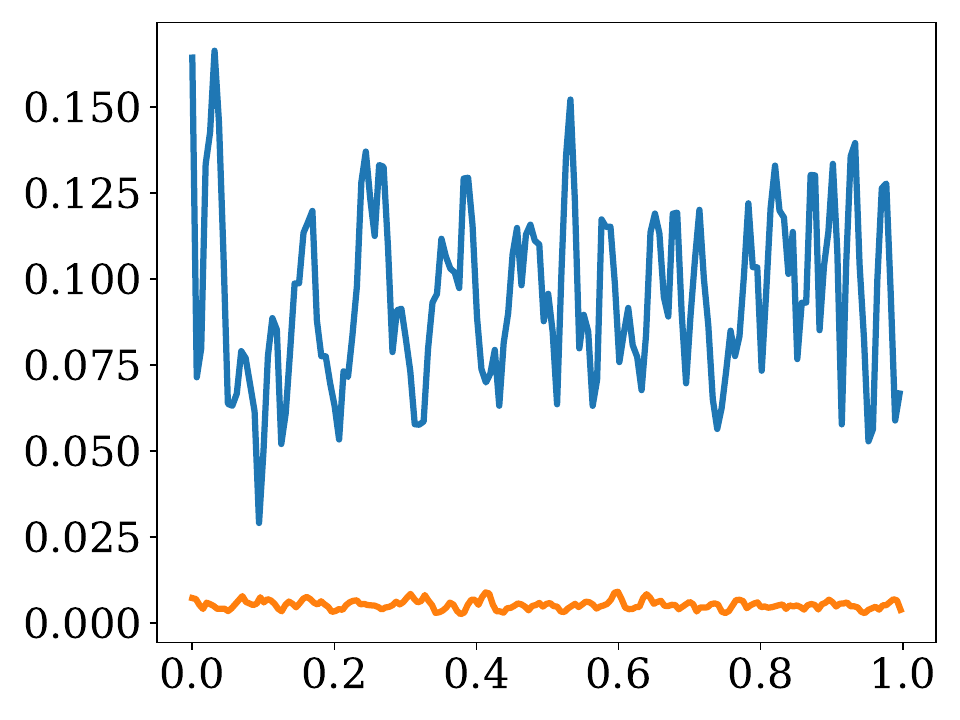}}
    \subfloat[Object Ratio]{
        \includegraphics[width=.235\linewidth]{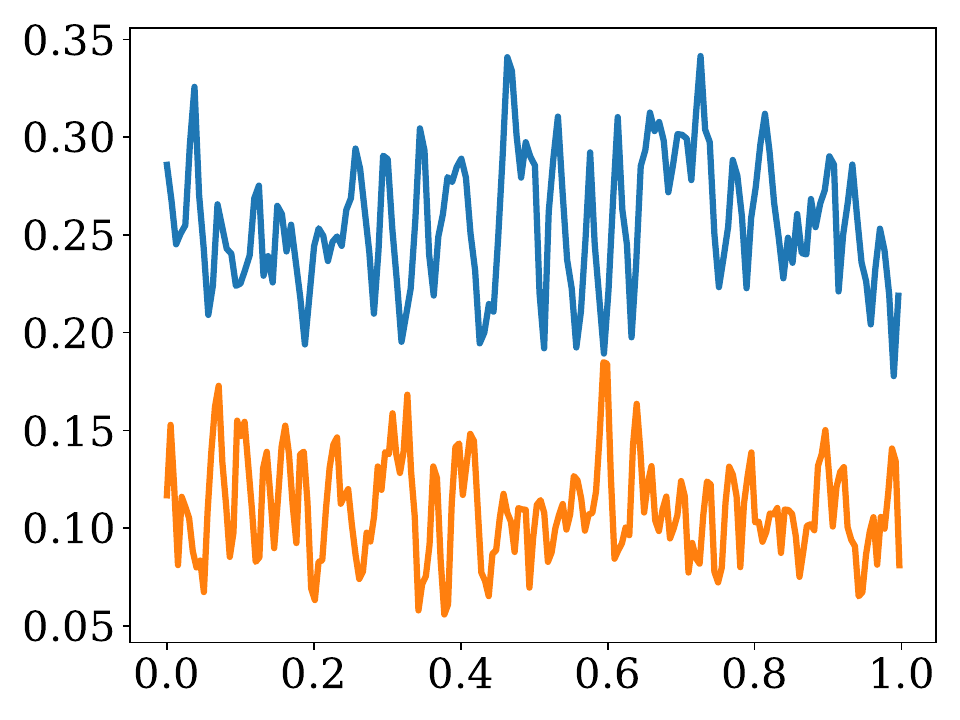}
    }
    \subfloat[Object Distance]{
        \includegraphics[width=.235\linewidth]{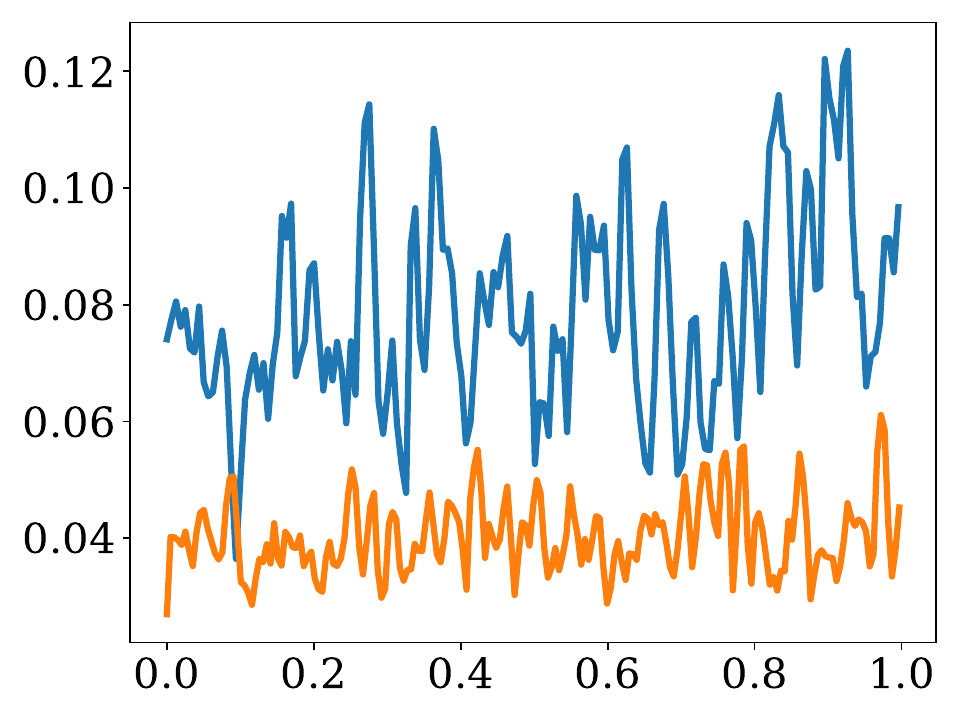}
    }
    \subfloat[Global Contrast]{
        \includegraphics[width=.235\linewidth]{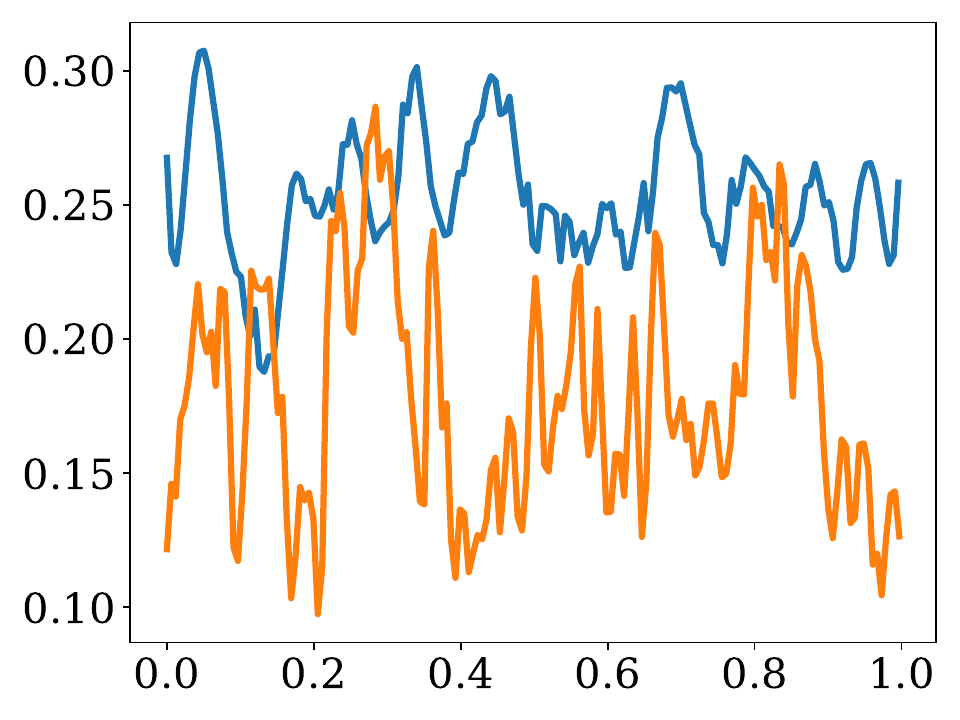}
    }
\caption{
    Comparisons of the attributes between \camsubset subset and \refsubset subset, \emph{i.e.}, Objects Area, Object Ratio, Object Distance, and Global Contrast. The results of the former are shown in orange, while the ones of the latter are shown in blue.}
    \label{fig:s2c7k-compare} 
\end{figure*}

\begin{figure}[t]
    \centering
    \footnotesize
    \begin{overpic}[width=0.91\linewidth]{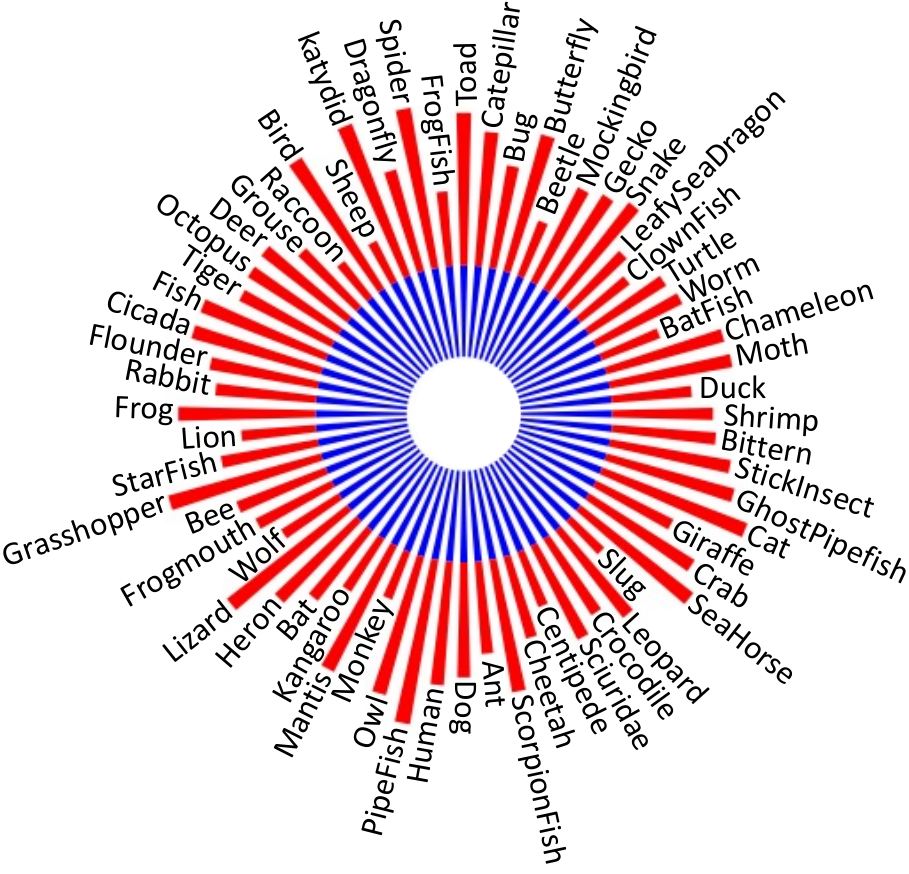}
        \put (45.5,49.5){\textbf{R2C7K}}
    \end{overpic}
   \caption{
        Taxonomic system and log number distribution of our \nameofdataset{} dataset.
        Note that the results of \camsubset subset are shown in red, and the ones of \refsubset subset are shown in blue.
   }
\label{fig:s2c7k-cate}
\end{figure}

\subsection{Data Statistics}
\myPara{Subset Comparisons.} 
\figref{fig:s2c7k-compare} presents four attribute comparisons between the images in \refsubset subset and \camsubset subset.
Specifically, the object area refers to the size of the objects in a given image, the object ratio is the proportion of objects in an image, the object distance is the distance from
the object center to the image center, and the global contrast is a metric to evaluate how challenging an object is to detect. 
It can be observed that the objects in the \refsubset subset are bigger than those objects in the \camsubset subset, and the images in the 
\refsubset subset contain more contrast cues. Therefore, the objects in the \refsubset subset are easier to be detected, and this form of referring information is suitable for the \nameofbenchmark{} research.

\begin{figure}[t]
    \centering
    \footnotesize
    \subfloat[\camsubset subset]{
        \includegraphics[width=.47\linewidth]{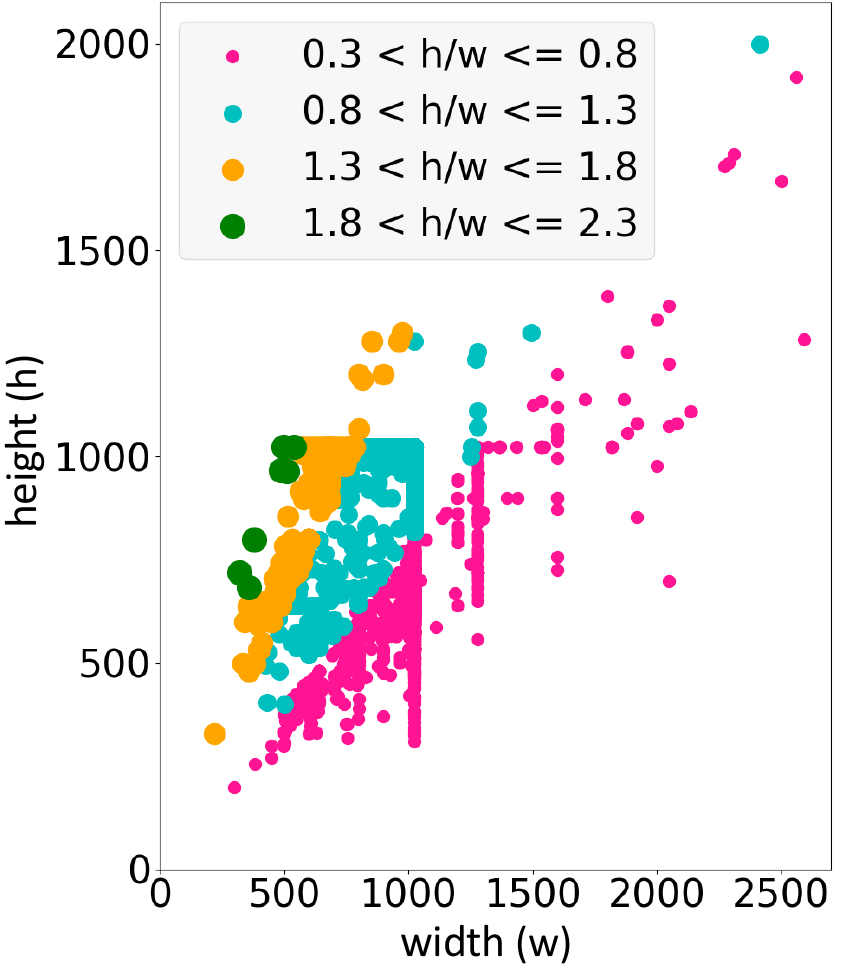}
        \label{fig:\camsubset-resol}
    }
    \subfloat[\refsubset subset]{
        \includegraphics[width=.47\linewidth]{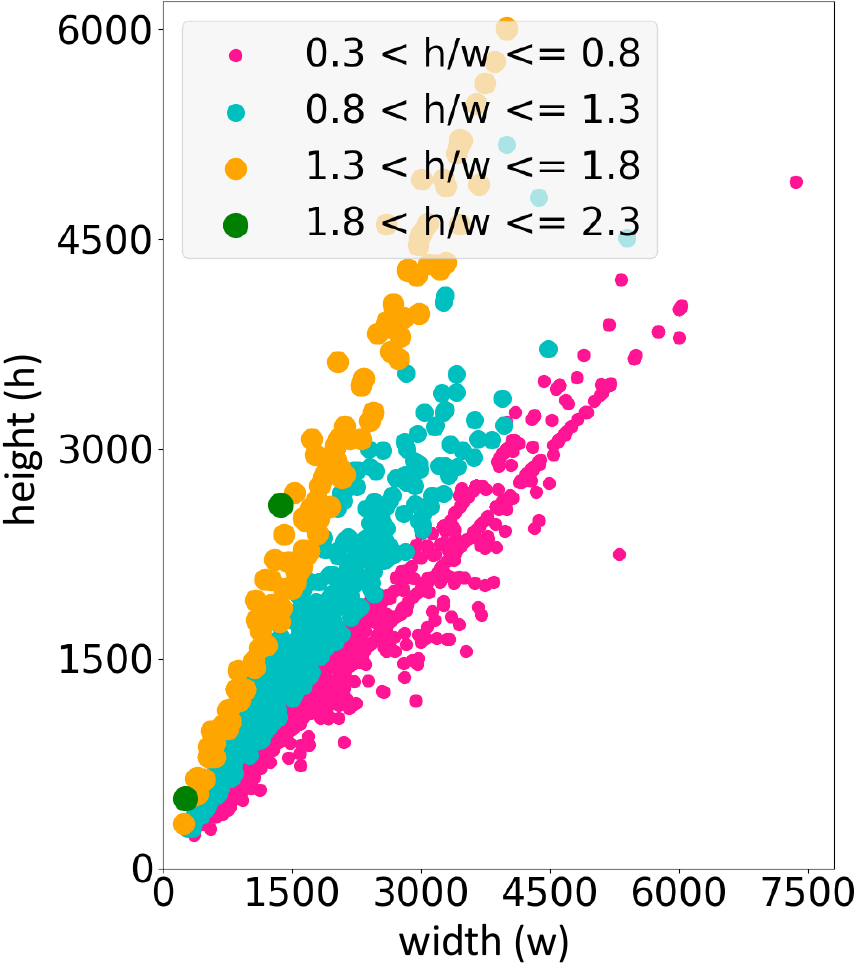}
        \label{fig:\refsubset-resol}
    }
   \caption{
        Image resolution distributions in \camsubset subset and \refsubset subset.
   }
\end{figure}

\begin{figure*}
\begin{center}
\includegraphics[width=\linewidth]{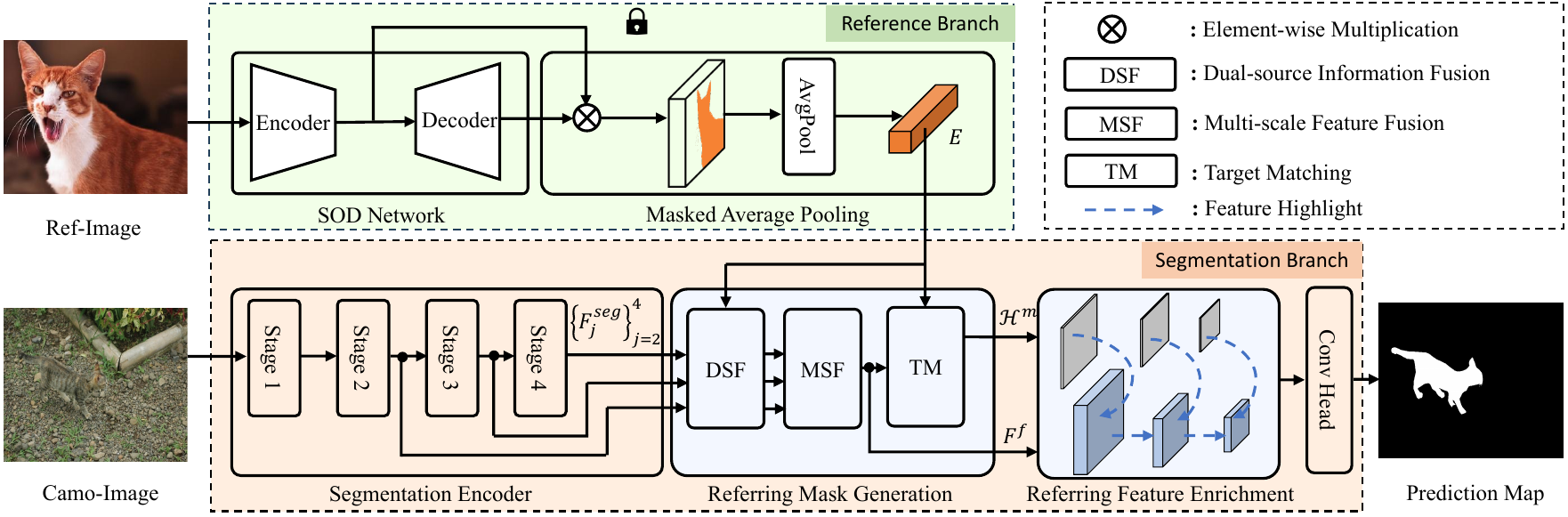}
\end{center}
\vspace{-10pt}
   \caption{
   Overall architecture of our \nameofmethod{} framework, which is composed of two branches, \emph{i.e.}, \refbranch{} in green and \mainbranch{} in orange. 
   In the \refbranch{}, the common representations of a specified object from images is obtained by masking and pooling the visual features with the foreground map generated by a SOD network.
   In the \mainbranch{}, the visual features from the last three layers of the encoder are employed to represent the given image.
   Then, these two kinds of feature representations are fused and compared in the well-designed RMG module to generate a mask prior, which is used to enrich the visual features among different scales to highlight the camouflaged targets in our RFE module. 
   Finally, the enriched features are fed into the decoder to generate the final segmentation map.
   }
\label{fig:s2cnet}
\end{figure*}

\arrayrulecolor{black}
\begin{table*}[htp!]
  \setlength\tabcolsep{6.2pt}
  \centering
  \arrayrulecolor{\nameofmodifycolor}
  \small
  \caption{\modifying{Comparisons of our \nameofdataset{} with previous COD datasets across a wide range of dimensions.
  Cate.: the number of categories.
  Camo-Img.: the number of images containing camouflaged objects;
  Ref-Img.: the number of referring images;
  Loc.: Location;
  Det.: Detection;
  Cls.: Classification;
  WS.: Weak Supervision;
  RefSeg.: Referring Object Segmentation.
  }}
  \label{tab:cod_dataset_comp}
  \begin{tabular}{lcccccccccc} 
        \toprule
        \multirow{2}{*}{\modifying{Datasets}} & \multicolumn{4}{c}{\makebox[0.13\textwidth][c]{\makecell{\modifying{Statistics}}}} & \multicolumn{4}{c}{\makecell{\modifying{Tasks}}}\\  \cmidrule(lr){2-5}  \cmidrule(lr){6-10}
        & \makecell{\modifying{Year}}& \makecell{\modifying{Cate.}}& \makecell{\modifying{Camo-Img.}}& \makecell{\modifying{Ref-Img.}} & \makecell{\modifying{Loc.}} & \makecell{\modifying{Det.}}&\makecell{\modifying{Cls.}}&\makecell{\modifying{WS.}} &\makecell{\modifying{RefSeg.}} \\ 
        \midrule
        \modifying{CHAMELEON~\cite{skurowski2018animal}} & \modifying{2018} & \modifying{N/A} & \modifying{76} & \modifying{N/A} & \modifying{\cmark} & \modifying{\cmark} & \modifying{\xmark} & \modifying{\xmark} & \modifying{\xmark} \\
        \modifying{CAMO~\cite{le2019anabranch}} & \modifying{2019} & \modifying{N/A} & \modifying{2500} & \modifying{N/A} & \modifying{\cmark} & \modifying{\cmark} & \modifying{\xmark} & \modifying{\xmark} & \modifying{\xmark} \\
        \modifying{COD10K~\cite{fan2022concealed}} & \modifying{2020} & \modifying{68} & \modifying{5066} & \modifying{N/A} & \modifying{\cmark} & \modifying{\cmark} & \modifying{\cmark} & \modifying{\cmark} & \modifying{\xmark} \\ 
        \modifying{NC4K~\cite{lv2021simultaneously}} &\modifying{2021} & \modifying{N/A} & \modifying{4121} & \modifying{N/A} & \modifying{\cmark} & \modifying{\cmark} & \modifying{\xmark} & \modifying{\xmark} & \modifying{\xmark} \\ 
        \textbf{\modifying{\nameofdataset{}~(Ours)}} & \modifying{2023} & \modifying{64} & \modifying{5015} & \modifying{1600} & \modifying{\cmark} & \modifying{\cmark} & \modifying{\cmark} & \modifying{\cmark} & \modifying{\cmark} \\ 
        \bottomrule
  \end{tabular}
\end{table*}



\myPara{Categories and Number.} 
The \nameofdataset{} dataset contains 6,615 samples covering 64 categories, where the \camsubset subset consists of 5,015 samples and the \refsubset subset has 1,600 samples. 
Note that, each category of this dataset contains a fixed number of referring images, namely 25, while the number of COD images in each category is unevenly distributed, as illustrated in \figref{fig:s2c7k-cate}.

\myPara{Resolution Distribution.} 
\figref{fig:\camsubset-resol} and \figref{fig:\refsubset-resol} shows the resolution distribution of the images in the \camsubset subset and \refsubset subset. 
As can be seen, these two subsets contain a large number of Full HD images, which can provide more details on the boundaries and textures of objects.

\myPara{Dataset Splits.} 
To facilitate the development of models for \nameofbenchmark{} research, we provide a referring
split for the \nameofdataset{} dataset. 
For the \refsubset subset, 20 samples are randomly selected from each category for training while the remaining 5 samples in each category are used for testing;
As for the \camsubset subset, the samples coming from the training set of COD10K are also utilized for training and the ones belonging to the test set are used for testing. 
The samples from NC4K are randomly assigned to the training and testing sets to ensure that each category in these two splits contains at least 6 samples.

\myPara{\modifying{Comparison with Existing Datasets.} }
\modifying{
We compare the proposed \nameofdataset{} dataset with the existing COD datasets, including CHAMELEON~\cite{skurowski2018animal}, CAMO~\cite{le2019anabranch}, COD10K~\cite{fan2022concealed}, and NC4K~\cite{lv2021simultaneously}, across a wide range of dimensions.
As shown in \tabref{tab:cod_dataset_comp}, the CHAMELEON dataset, collected from Google search engine with a keyword, \ie ``concealed animal'', only contains 76 images.
The CAMO dataset has 2,500 images containing camouflaged objects across 8 categories.
The COD10K and NC4K datasets have a higher number of images containing camouflaged objects, with 5,066 and 4,121 images respectively.
These datasets greatly promote the development of camouflage object detection.
Nevertheless, they still have limitations when it comes to a real scene
with multiple types of camouflaged objects, hindering their practical application.
In contrast, our \nameofdataset{} dataset has a large number of simple scene images (1,600) covering multiple object categories (64) so that users can choose them as references according to their needs to search for specified camouflaged objects in complex scenes (5,015).
As a result, it has the potential to benefit more tasks.}

\section{Proposed Framework}

In this section, we first briefly describe the definition of our \nameofbenchmark{} task in \secref{subsec:task-des}. Then, we introduce the overall architecture of our \textbf{\nameofmethod{}} in \secref{subsec:overall}. Next, we present the details of the \genmoudule{} (\textbf{RMG}) module and the \enrmoudule{} (\textbf{RFE}) module in \secref{subsec:rmg} and \secref{subsec:rfe}, respectively.

\subsection{Task Description} \label{subsec:task-des}

Our \nameofbenchmark{} aims at segmenting the specified camouflaged objects under the guidance of several referring images.
Different from the standard COD that detects all the camouflage objects, our \nameofbenchmark{} attempts to identify the specific objects by matching the targets provided by the references.
In particular, the input of a \nameofbenchmark{} system is composed of two parts, 
where the first part is an image with camouflaged objects, termed as $I^{camo} \in \mathbb{R}^{3 \times H \times W}$, while the other part is a few referring images with salient target objects, denoted as $I^{ref} = \{I^{ref}_{i}\}_{i=1}^{K}, I^{ref}_{i} \in \mathbb{R}^{3 \times H \times W}$, 
where $H$ and $W$ denote the height and width of the given images, and $K$ represents the number of referring images.
Note that $I^{camo}$ comes from the \camsubset subset of \nameofdataset{}, and it contains camouflaged objects in a specified category \emph{c}.
Meanwhile, $I^{ref}_{i}$ is sampled from the \refsubset subset, and its salient objects are in category \emph{c}.
The output of \nameofbenchmark{} is a binary mask $M^{seg}$ for the camouflaged objects of category \emph{c} in $I^{camo}$.

\subsection{Overall Architecture} \label{subsec:overall}
\figref{fig:s2cnet} illustrates the overall architecture of our proposed \nameofmethod{}. As can be seen, this framework consists of two branches, \emph{i.e.}, the \refbranch{} and the \mainbranch{}, which will be described in detail.

\myPara{Reference Branch.} 
The pipeline of extracting common representation from referring images is cascaded by a SOD network based on the encoder-decoder structure and a masked average pooling (MAP) function.
In particular, the former is used to generate the visual features and foreground predictions of the referring images, and the latter one is employed to filter out irrelevant information in the referring images. 
By default, the pre-trained ICON~\cite{zhuge2022salient} model with ResNet-50~\cite{he2016deep} backbone is selected as our SOD network.

Given $K$ referring images with salient target objects and spatial size $H \times W$, the visual features with spatial size $\frac{H}{32} \times \frac{W}{32}$ and foreground maps with spatial size $H \times W$ are first extracted from the encoder and decoder of the SOD network.
We denote them as $\{F^{ref}_{k}\}_{k=1}^{K}$ and $\{M^{ref}_{k}\}_{k=1}^{K}$, respectively. 
Then, $\{F^{ref}_{k}\}_{k=1}^{K}$ and $\{M^{ref}_{k}\}_{k=1}^{K}$ are fed into the MAP function to calculate representations associated with the foreground objects, denoted as $F^{obj}_{k} \in \mathbb{R}^{c_{d} \times 1 \times 1}$. 
This calculation can be formulated as:
\begin{equation} 
   F^{obj}_{k} = \mathcal{F}\mathrm{_{conv1\times1}} \bigg(
            \frac{\sum_{2d} (\mathcal{F}\mathrm{_{down}}(M^{ref}_{k}) \otimes F^{ref}_{k})}{\sum_{2d}(F^{ref}_{k})}
   \bigg), 
  \label{eq:overall-map}
\end{equation}
where $\otimes$ is the element-wise multiplication operation, 
$\mathcal{F}\mathrm{_{down}}(\cdot)$ is a bilinear downsampling operation for shape matching,
$\sum_{2d}(\cdot)$ accumulates the feature values along the spatial dimension, 
and $\mathcal{F}\mathrm{_{conv1\times1}}(\cdot)$ is a $1 \times 1$ convolution that transforms the channel of its input to $c_{d}$ to achieve a better trade-off between efficiency and performance. 
Finally, we abstract the common representations of target objects in the embedding space, denoted as $E \in \mathbb{R}^{c_{d} \times 1 \times 1}$, by averaging these object representations.

\myPara{Segmentation Branch.}
Our \mainbranch{} is built upon an encoder-decoder structure as well, which has been widely used in COD research. 
Note that as the goal of this paper is to provide a new scheme for directionally segmenting the camouflaged objects, we do not put too much effort on the architectural design.
In fact, we find that under the proposed scheme, even a simple segmentation network can still perform well.
Thus, ResNet-50~\cite{he2016deep} is adopted as the encoder, and the features from its last three layers are selected as visual representations, following previous works~\cite{fan2022concealed,ji2023deep}.
The decoder is a convolutional head consisting of two convolutional layers for the identification of camouflaged objects.
We also propose two new modules, namely referring mask generation (RMG) and referring feature enrichment (RFE), 
which are added between the encoder and the decoder, to take advantage of the common representations from the reference branch
for the explicit segmentation of camouflaged targets.

Given an image with camouflaged objects and spatial size $H \times W$, we first extract multi-scale features from the last three stages from the encoder. The channels of these features are converted to $c_{d}$ using $1 \times 1$ convolutions.
We denote these features as
$\{F^{seg}_{j}\}_{j=2}^{4}$, where $F^{seg}_{j} \in \mathbb{R}^{c_{d} \times \frac{H}{2^{j+1}} \times \frac{W}{2^{j+1}}}$.
Then, the resulting features are fed into the RMG module together with the common representations from the reference branch to generate fusion features and a referring mask, which are denoted as $F^{f} \in \mathbb{R}^{c_{d} \times \frac{H}{8} \times \frac{W}{8}}$ and $\mathcal{H}^{m} \in \mathbb{R}^{1 \times \frac{H}{8} \times \frac{W}{8}}$. 
This process can be defined as:
\begin{equation} 
   F^{f}, \mathcal{H}^{m} = \mathrm{RMG}(\{F^{seg}_{j}\}_{j=2}^{4}, E).
  \label{eq:overall-fusion}
\end{equation}
Note that $\mathcal{H}^{m}$ is a heatmap mask, in which higher scores mean the corresponding positions are more associated with the common representations, and vice versa.

Next, the fusion features are enriched at multiple scales with the guidance of the prior mask to highlight the camouflaged targets in the RFE module, which is defined as:
\begin{equation} 
   F^{enr} = \mathrm{RFE}(F^{f}, \mathcal{H}^{m}).
  \label{eq:overall-guidance}
\end{equation}
Finally, the resulting features are fed into the decoder to generate the segmentation map $M^{seg} \in \mathbb{R}^{1 \times H \times W}$.

\subsection{\genmoudule{}} \label{subsec:rmg}
In order to identify the camouflaged objects accurately from their highly similar surroundings according to the image guidance, a pixel-level comparison between the common representations of target objects and the visual features should be performed. 
However, the appearance of a camouflaged object may be significantly different from that of the referring objects even if they belong to the same category. 
In addition, the common representations and the visual features come from different information sources. 
Such a large information difference may interfere with the comparison process, making the localization process of the camouflaged targets difficult. 

\begin{figure}[t]
    \centering
    \includegraphics[width=.92\linewidth]{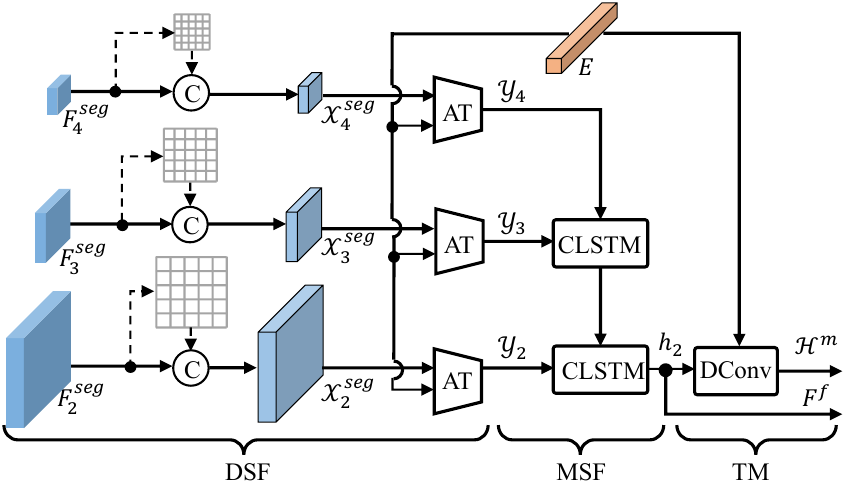}
   \caption{
        Structural details of the proposed Referring Mask Generation (RMG) module.
        Note that `C' is the concatenation operation, `AT' is the affine transformation, `CLSTM' is the convolutional long short-term memory, and 'Dconv' is the dynamic convolution.
   }
   \label{fig:rmg}
\end{figure}

Motivated by recent works on multimodal fusion~\cite{huang2021look, li2018referring}, we solve this issue by performing a dual-source information fusion (DSF) between the common representations and the visual features, as illustrated in \figref{fig:rmg}. 
To promote the interaction of these two kinds of information, the spatial position information is first injected into the visual features. 
Specifically, each position of the visual features is concatenated with an 8d embedding vector akin to the implementation of ~\cite{yang2019fast}.
We denote the visual features equipped with spatial information as $\{x^{seg}_{j}\}_{j=2}^{4}$, where $x^{seg}_{j} \in \mathbb{R}^{(c_{d}+8) \times \frac{H}{2^{j+1}} \times \frac{W}{2^{j+1}}}$. 
Then, we apply an affine transformation on the visual features under the guidance of the common representations. 
In particular, two linear layers are utilized to map the common representations to two coefficient vectors, which are applied on the visual features followed by convolution and ReLU operations. The obtained results are denoted as $\{y_{j}\}_{j=2}^{4}$, where $y_{j} \in \mathbb{R}^{c_{d} \times \frac{H}{2^{j+1}} \times \frac{W}{2^{j+1}}}$. 
This process is defined as follows:
\begin{equation} 
   y_{j} = \mathcal{F}_{relu}(\mathcal{F}_{conv3\times3}(\mathcal{F}_{relu}(\mathcal{\gamma}_{j} \otimes x^{c}_{j}\oplus \beta_{j}))),
  \label{eq:dif-film2}
\end{equation}
\begin{equation} 
   \mathcal{\gamma}_{j} = \mathcal{F}_{mlp1}(E), \mathcal{\beta}_{j} = \mathcal{F}_{mlp2}(E),
  \label{eq:dif-film1}
\end{equation}
where $\otimes$ and $\oplus$ are the element-wise multiplication and addition respectively, 
$\mathcal{F}_{conv3\times3}(\cdot)$ is a $3 \times 3$ convolution for channel recovery,
$\mathcal{\gamma}_{j}$ and $\mathcal{\beta}_{j}$ are the two coefficient vectors, and
$\mathcal{F}_{mlp1}(\cdot)$ and $\mathcal{F}_{mlp2}(\cdot)$ denote the two linear layers.

\begin{figure}[t]
    \centering
    \includegraphics[width=.97\linewidth]{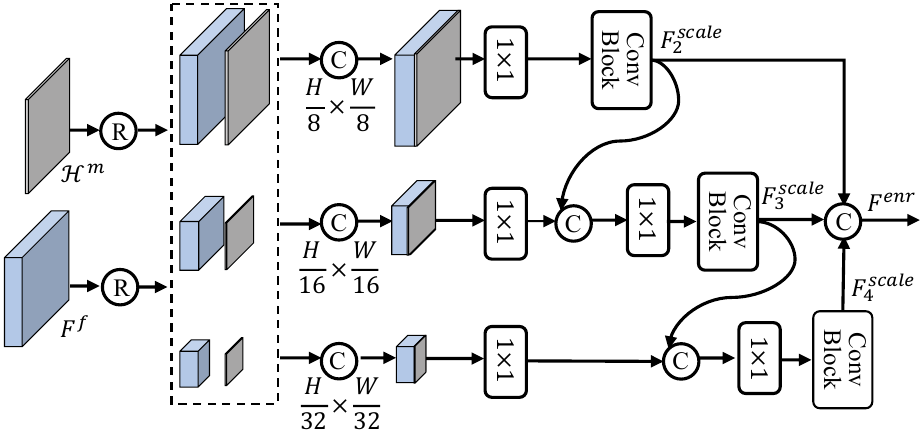}
   \caption{
        Structural details of our Referring Feature Enrichment (RFE) module.
        Note that `R' is the resize operation, `$1\times1$' is a convolutional layer with kernel size 1, and `Conv Block' is composed of two convolutional layers with kernel size 3.
   }
   \label{fig:rfe}
\end{figure}

Furthermore, multi-scale fusions (MSF) are performed on the dual-source fusion features along the up-bottom path via convolutional LSTM~\cite{shi2015convolutional} to enhance the robustness of our method to targets with different scales. This process can be formulated as follows:
\begin{equation} 
    h_{j}, c_{j} = \mathcal{F}_{clstm}(y_{j}, [\mathcal{F}_{up}(h_{j+1}), \mathcal{F}_{up}(c_{j+1})])),
  \label{eq:dif-clstm}
\end{equation}
where $\mathcal{F}\mathrm{_{up}}(\cdot)$ is a bilinear upsampling operation for shape matching and $\mathcal{F}_{clstm}$ denotes a cell of convolutional LSTM. 
Note that the initial state $h_{4}=c_{4}=y_{4}$, 
and the last hidden state, \emph{i.e.}, $h_{2} \in \mathbb{R}^{c_{d} \times \frac{H}{8} \times \frac{H}{8}}$, is used as the fused feature $F^{f}$.

\arrayrulecolor{black}
\begin{table*}[tp]
\setlength\tabcolsep{4pt}
\centering
\arrayrulecolor{\nameofmodifycolor}
\caption{ \modifying{
Comparisons of popular COD models with their \nameofbenchmark{} counterparts. 
All models are evaluated on a NVIDIA RTX 3090 GPU.
`R-50': ResNet-50~\cite{he2016deep}, `E-B4': EfficientNet-B4~\cite{tan2019efficientnet}, `R2-50': Res2Net-50~\cite{gao2019res2net}, `$\rm R^{3}$-50': Triple ResNet-50~\cite{pang2022zoom}, `Swin-S': SwinTransformer-S~\cite{liu2021swin},
`-\salsuffix': models with image references composed of salient objects, 
`Attribute': attributes of each network, `Single-obj': scenes of a single camouflaged object, `Multi-obj': scenes of multiple camouflaged objects, `Overall': all scenes containing camouflaged objects,
`$\uparrow$': the higher the better, `$\downarrow$': the lower the better.}}
\resizebox{\linewidth}{!}{
\begin{tabular}{lcccccccccccccccc}
\toprule
\multirow{2}{*}{Models} & \multicolumn{4}{c}{\makebox[0.22\textwidth][c]{\makecell{Attribute}}} & \multicolumn{4}{l}{\makebox[0.18\textwidth][c]{\makecell{Overall}}} & \multicolumn{4}{l}{  \makebox[0.18\textwidth][c]{\makecell{Single-obj}}} & \multicolumn{4}{l}{  \makebox[0.18\textwidth][c]{\makecell{Multi-obj}}}\\
\cmidrule(lr){2-5}
\cmidrule(lr){6-9}
\cmidrule(lr){10-13}
\cmidrule(lr){14-17}
& Backbone & Params (M) & Macs (G) & Speed (FPS) & \makecell{S$_m$ $\uparrow$} &\makecell{$\alpha $E $\uparrow$}  &\makecell{$w$F  $\uparrow$} &\makecell{M$\downarrow$} & \makecell{S$_m$ $\uparrow$} &\makecell{$\alpha $E $\uparrow$}  &\makecell{$w$F  $\uparrow$} &\makecell{M$\downarrow$} & \makecell{S$_m$ $\uparrow$} &\makecell{$\alpha $E $\uparrow$}  &\makecell{$w$F  $\uparrow$} &\makecell{M$\downarrow$}\\ 
\midrule
Baseline                                     & R-50           & 25.97 & 21.02 & 185.74 & 0.772 & 0.847 & 0.604 & 0.044 & 0.777 & 0.847 & 0.611 & 0.043 & 0.711 & 0.849 & 0.531 & 0.054\\
\rowcolor[HTML]{EFEFEF}
\textbf{\nameofmethod{}}                       & R-50           & 27.15 & 23.23 & 151.47 & 0.805 & 0.879 & 0.669 & 0.036 & 0.810 & 0.880 & 0.674 & 0.035 & 0.747 & 0.872 & 0.602 & 0.046\\
\midrule
$\rm PFNet_{2021}$~\cite{mei2021camouflaged} & R-50          & 48.55 & 52.99 & 80.33  & 0.791 & 0.876 & 0.651 & 0.040 & 0.795 & 0.876 & 0.656 & 0.039 & 0.74 & 0.868 & 0.594 & 0.051\\
\rowcolor[HTML]{EFEFEF}
\textbf{PFNet-\salsuffix}                        & R-50          & 57.58 & 59.59 & 72.48  & 0.811 & 0.885 & 0.687 & 0.036 & 0.815 & 0.886 & 0.691 & 0.035 & 0.764 & 0.873 & 0.632 & 0.045\\
\midrule
$\rm PreyNet_{2022}$~\cite{zhang2022preynet} & R-50          & 38.53 & 116.01 & 59.78 & 0.806 & 0.890 & 0.690 & 0.034 & 0.811 & 0.892 & 0.696 & 0.033 & 0.749 & 0.878 & 0.618 & 0.042\\
\rowcolor[HTML]{EFEFEF}
\textbf{PreyNet-\salsuffix}                        & R-50          & 38.70 & 117.60 & 57.04  & 0.817 & 0.900 & 0.704 & 0.032 & 0.822 & 0.900 & 0.709 & 0.032 & 0.763 & 0.898 & 0.645 & 0.041\\
\midrule
$\rm SINetV2_{2022}$~\cite{fan2022concealed} & R2-50          & 26.98 & 24.48 & 98.12  & 0.813 & 0.874 & 0.678 & 0.036 & 0.818 & 0.874 & 0.684 & 0.035 & 0.763 & 0.864 & 0.615 & 0.045\\
\rowcolor[HTML]{EFEFEF}
\textbf{SINetV2-\salsuffix }                        & R2-50          & 27.70 & 26.01 & 86.60  & 0.823 & 0.888 & 0.700 & 0.033 & 0.828 & 0.889 & 0.705 & 0.032 & 0.771 & 0.874 & 0.634 & 0.043\\
\midrule
$\rm BSANet_{2022}$~\cite{zhu2022can} & R2-50           & 32.59 & 59.29  &  71.75 & 0.818 & 0.893 & 0.702 & 0.034 & 0.823 & 0.895 & 0.707 & 0.033 & 0.766 & 0.873 & 0.643 & 0.041\\
\rowcolor[HTML]{EFEFEF}
\textbf{BSANet-\salsuffix}                           & R2-50           & 33.07 & 66.08  &  67.18 & 0.830 & 0.912 & 0.727 & 0.030 & 0.827 & 0.913 & 0.733 & 0.030 & 0.774 & 0.895 & 0.655 & 0.039\\
\midrule
$\rm BGNet_{2022}$~\cite{sun2022boundary} & R2-50 & 79.85 & 116.76 & 66.29 & 0.818 & 0.901 & 0.679 & 0.036 & 0.822 & 0.901 & 0.683 & 0.035 & 0.775 & 0.886 & 0.626 & 0.044\\
\rowcolor[HTML]{EFEFEF}
\textbf{BGNet-\salsuffix}  & R2-50 & 151.06 & 171.03 & 50.69 & 0.840 & 0.909 & 0.738 & 0.029 & 0.844 & 0.910 & 0.742 & 0.029 & 0.792 & 0.887 & 0.679 & 0.036\\
\midrule
$\rm ZoomNet_{2022}$~\cite{pang2022zoom}     & $\rm R^{3}$-50 & 32.38 & 203.50 & 22.89 & 0.813 & 0.884 & 0.688 & 0.032 & 0.818 & 0.885 & 0.695 & 0.031 & 0.747 & 0.870 & 0.605 & 0.042\\
\rowcolor[HTML]{EFEFEF}
\textbf{ZoomNet-\salsuffix}& $\rm R^{3}$-50 & 33.30 & 218.24 & 20.82 &0.834&0.886&0.720&0.029&0.839&0.887&0.726&0.029&0.781&0.876&0.652&0.038\\
\midrule
$\rm DGNet_{2023}$~\cite{ji2023deep} & E-B4           & 19.22 & 5.53  &  110.57 & 0.816 & 0.883 & 0.684 & 0.034 & 0.826 & 0.885 & 0.700 & 0.032 & 0.744 & 0.873 & 0.588 & 0.047\\
\rowcolor[HTML]{EFEFEF}
\textbf{DGNet-\salsuffix}                           & E-B4           & 20.10 & 7.24  &  95.06 & 0.821 & 0.891 & 0.696 & 0.032 & 0.827 & 0.890 & 0.703 & 0.031 & 0.748 & 0.879 & 0.607 & 0.045\\
\midrule
\modifying{$\rm VSCode_{2024}$~\cite{luo2024vscode}}                           &   \modifying{Swin-S}        & \modifying{74.72}  & \modifying{59.81}  & \modifying{76.81} & \modifying{0.819} & \modifying{0.879} & \modifying{0.702} & \modifying{0.033} & \modifying{0.825}  & \modifying{0.880}  & \modifying{0.706} & \modifying{0.032} & \modifying{0.750} & \modifying{0.868} & \modifying{0.651} & \modifying{0.043} \\
\rowcolor[HTML]{EFEFEF}
\textbf{\modifying{VSCode-\salsuffix}}                           &   \modifying{Swin-S}        & \modifying{76.63}  & \modifying{64.28}  & \modifying{65.26} &  \modifying{0.832} & \modifying{0.891} & \modifying{0.714} & \modifying{0.030} &  \modifying{0.838} & \modifying{0.892} &  \modifying{0.718} &  \modifying{0.029} & \modifying{0.766}  &  \modifying{0.880} & \modifying{0.662} & \modifying{0.041} \\
\midrule
\modifying{$\rm ZoomNext_{2024}$~\cite{pang2024zoomnext}}                           &   \modifying{$\rm R^{3}$-50}      & \modifying{28.46}  & \modifying{185.79}  & \modifying{66.29}  &  \modifying{0.838}  & \modifying{0.897} & \modifying{0.742} & \modifying{0.032} & \modifying{0.843}  & \modifying{0.898} & \modifying{0.750} & \modifying{0.031} & \modifying{0.777} & \modifying{0.880} & \modifying{0.655} & \modifying{0.040} \\
\rowcolor[HTML]{EFEFEF}
\textbf{\modifying{ZoomNext-\salsuffix}}                           &   \modifying{$\rm R^{3}$-50}        &\modifying{30.32} &\modifying{197.43}&\modifying{52.81} &  \modifying{0.850} & \modifying{0.909} &  \modifying{0.755}  & \modifying{0.027} & \modifying{0.859}  & \modifying{0.910} & \modifying{0.762}  & \modifying{0.026}  &  \modifying{0.788}  &  \modifying{0.892} & \modifying{0.675} &  \modifying{0.037} \\
\bottomrule
\end{tabular}}
\vspace{1pt}
\label{tab:adaptation}
\end{table*}
\arrayrulecolor{black}

\begin{figure*}[t]
	\centering
        \footnotesize
        \subfloat[\modifying{PR curves}]{
             \includegraphics[width=0.99\linewidth]{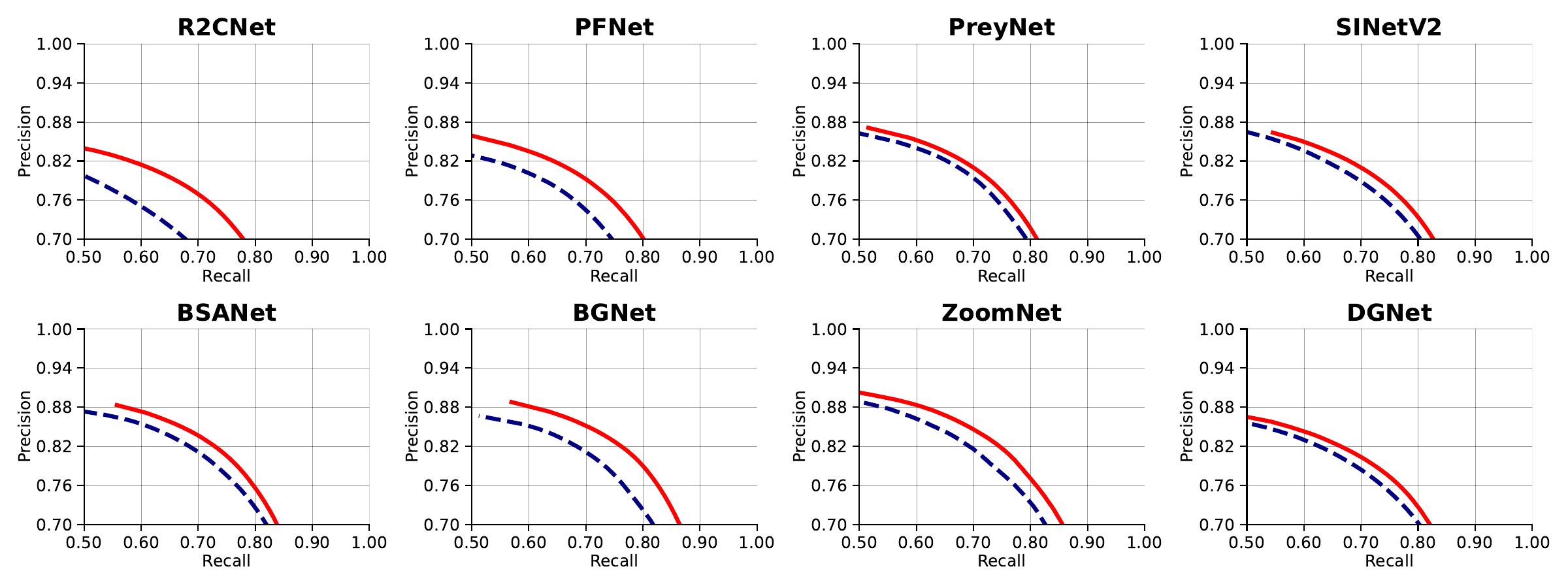}
        } \\
        \subfloat[\modifying{$F_\beta$ curves}]{
            \includegraphics[width=0.99\linewidth]{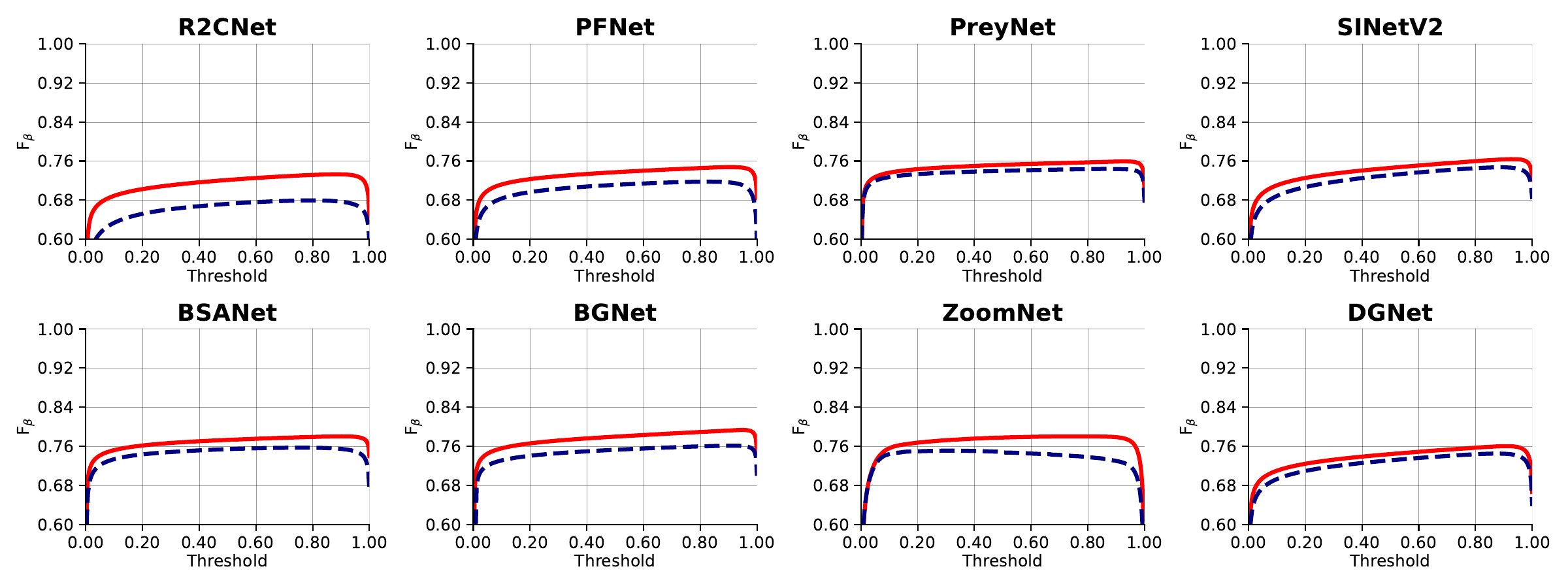}
        }
        \caption{
        \modifying{
        PR and $F_{\beta}$ curves of the COD methods and their \nameofbenchmark{} variants. 
        The results of standard COD methods and their Ref-COD counterparts
        are illustrated via blue dashed lines and red solid lines respectively.}
        }
    \label{fig:curves} 
\end{figure*}

Finally, the common representations are compared with the representations of each position of the fusion features to produce a referring mask, namely $\mathcal{H}^{m}$. 
Inspired by~\cite{jing2021locate,liu2022dynamic}, this target-matching (TM) process is realized through a dynamic convolution operation. 
In particular, the common representations and the fusion features are utilized as reference-guided dynamic kernel and input, respectively.

\subsection{\enrmoudule{}} \label{subsec:rfe}

Given the referring mask generated above, we design a \enrmoudule{} (RFE) module to enrich the visual features at different scales, as shown in Fig~\ref{fig:rfe}. 

To be specific, the prior mask and the fusion features are first resized to the shapes of the aforementioned three features, \emph{i.e.}, $\{\frac{H}{2^{j+1}} \times \frac{W}{2^{j+1}} \}_{j=2}^{4}$, respectively. 
Then, the resized masks and features at the same scale are fused by concatenation, and the output features of different scales, termed as $\{F^{scale}_{j}\}_{j=2}^{4}$, where $F^{scale}_{j} \in \mathbb{R}^{c_{d} \times \frac{H}{2^{j+1}} \times \frac{W}{2^{j+1}}}$, are also concatenated together to enrich the capability of identifying camouflaged objects. 
We denote the enriched feature as $F^{enr} \in \mathbb{R}^{c_{d} \times \frac{H}{8} \times \frac{W}{8}}$. 
As mentioned in PFENet~\cite{tian2020prior}, tiny objects may become unclear in the down-sampled feature maps. 
Thus, we build a similar path that crosses scales from the finer feature to the coarse one to achieve the interaction between them. 
In addition, we also apply supervision on $\{F^{scale}_{j}\}_{j=2}^{4}$ to make the yielded feature $F^{enr}$ more robust, and the corresponding maps is denoted as $\{M^{scale}_{j}\}_{j=2}^{4}$, where $M^{scale}_{j} \in \mathbb{R}^{1 \times H \times W}$.

\section{Experiments}
In this section, we first introduce the experiment settings of this paper, \emph{i.e.}, training \& testing protocols, hyperparameter details, and evaluation metrics. 
Then, we conduct quantitative comparisons between the \nameofbenchmark{} methods and the corresponding COD methods. Particularly, we compare the proposed \nameofmethod{} with the baseline model on different camouflaged scenes. we also apply the design of \nameofbenchmark{} on 7 recent \sOTa{} COD methods to verify its generality. 
Next, we report ablative study results to investigate the effectiveness of each component and our design choice.
Finally, we carry out visualization comparisons to gain more intuitive insight into our \nameofbenchmark{}.

\subsection{Experiment Setup} \label{subsec:exp-setting}
\myPara{Training \& Testing Protocols.}
Considering that the goal of \nameofbenchmark{} is to generate a binary foreground map closing to the annotation, the structure loss~\cite{wei2020f3net} consisting of BCE loss and IoU loss, which has been widely used in many binary segmentation tasks, can be adopted as the optimization objectives of the \nameofbenchmark{} models. 
Specifically, this function can be formulated as: 
\begin{equation} 
   \mathcal{L}(P, G) = \sum_{i=1}^{4} \mathcal{L}_{bce} (P_{i}, G)+\mathcal{L}_{iou} (P_{i}, G), 
  \label{eq:loss}
\end{equation}
where $P = \{M^{scale}_{2}, M^{scale}_{3}, M^{scale}_{4}, M^{seg}\}$ represent the predictions generated by our model, and $G$ refers to the ground truth annotation.

\myPara{Hyperparameter Details.}
By default, the number of referring images in our \nameofbenchmark{} is set to 5. 
During the training stage, the parameters of the foreground prediction network are frozen, the batch size is set to 32, the optimizer we adopt is Adam~\cite{kingma2014adam}, and the learning rate is initialized to 5e-4 and gradually decays according to the cosine annealing method~\cite{loshchilov2016sgdr}. 
During inference, all images are first resized to $352 \times 352$ and then fed into the trained \nameofmethod{} to generate the final prediction without any post-processing operations. 
All experiments are implemented in PyTorch~\cite{paszke2019pytorch}. 

\myPara{Metrics.}
Following the standard protocol in COD, we adopt four common metrics for evaluation, including mean absolute error (M)~\cite{perazzi2012saliency}, structure-measure (S$_m$)~\cite{fan2017structure}, adaptive E-measure ($\alpha$E)~\cite{fan2018enhanced} and weighted F-measure ($w$F)~\cite{margolin2014evaluate}. 
To be specific, M measures the absolute difference between the predicted mask and ground-truth (GT); S$_m$ measures the region-aware and object-aware structural similarity between the predicted map and GT; $\alpha$E measures the element-wise and image-level similarity. $w$F is an exhaustive measure of both recall and precision between the prediction and GT.
In addition, we draw the precision-recall (PR) curves and $F_\beta$-threshold ($F_\beta$) curves for further comparison. 

\arrayrulecolor{black}

\begin{table*}[htp!]
  \setlength\tabcolsep{6pt}
  \centering
  \small
  \arrayrulecolor{\nameofmodifycolor}
  \caption{\modifying{Application of the paradigm of our \nameofbenchmark{} to recent published salient object detection models. 
  }}
  \color{\nameofmodifycolor}
  \label{tab:latest_sod_comp}
  \begin{tabular}{l|c|cccc|ccc} \toprule
    Methods & Publication & S$_m \uparrow$ & $\alpha $E $\uparrow$ & $w$F $\uparrow$ &M$\downarrow$ & Params (M) & MACs (G) & FPS \\ \midrule
    VST++ & \multirow{2}{*}{TPAMI2024} & 0.846  &0.872 & 0.795 &0.063  & 53.6 & 28.4 & 185.5 \\
    VST++-Ref & &0.858 & 0.885&0.810  & 0.057   &55.2 &34.1 & 166.9 \\
    \midrule
    VSCode & \multirow{2}{*}{CVPR2024} & 0.861  & 0.893 & 0.811 & 0.054 & 74.7 & 59.8 & 144.6 \\ 
    VSCode-Ref && 0.874 & 0.906 &0.820 &0.052    & 76.6&64.3 &128.3 \\ \bottomrule
  \end{tabular}
\end{table*}

\subsection{Quantitative Evaluation} \label{subsec:quan}
\myPara{Comparison with Baselines.} 
To verify the effectiveness of our \nameofmethod{}, we first compare it with its baseline variant, which is a standard COD model based on encoder-decoder architecture.
In this baseline model, the last three features from its encoder are fused along the up-bottom path, following the fashion in FPN, and the fused feature is fed to its decoder to directly segment camouflaged objects without any reference.
As shown in \tabref{tab:adaptation}, our \nameofmethod{} surpasses the baseline model by a large margin across all metrics in the overall scene (\emph{i.e.}, test split) of \nameofdataset{}. 
To evaluate the performance of the model more comprehensively, we further divide the overall scene into two groups, namely the scene of single camouflaged objects and the scene of multiple camouflaged objects.
It can be observed that our \nameofmethod{} still outperforms its COD baseline in all settings, which demonstrates that the referring images contribute to the identification of the camouflaged objects from the confusing scenes.

\myPara{Application to Existing COD Methods.} 
To verify the generality of our \nameofbenchmark{} design, we also apply this idea to existing COD methods.
These methods are chosen according to three criteria: a) recently published, b) representative, and c) with open source code. 
In particular, the methods to participate in the adaptation here include: 
PFNet~\cite{mei2021camouflaged}, PreyNet~\cite{zhang2022preynet}, SINetV2~\cite{fan2022concealed}, BSANet~\cite{zhu2022can}, 
BGNet~\cite{sun2022boundary}, ZoomNet~\cite{pang2022zoom},  DGNet~\cite{ji2023deep},
\modifying{VSCode~\cite{luo2024vscode},
and ZoomNext~\cite{pang2024zoomnext}}. 
As shown from the third to the last row of \tabref{tab:adaptation} and \figref{fig:curves}, the \nameofbenchmark{} variants of these models achieve better performance than their original versions in all metrics. 
It should be noted that our lightweight baseline model even achieves a performance comparable to recent COD methods after incorporating reference images.
This indicates that our method does not rely on powerful segmentation networks.
The introduction of the reference branch can help significantly improve the performance of a COD model even though it is not strong, reflecting the effectiveness of our \nameofbenchmark{} idea.
\modifying{
Moreover, the relative order of metrics values for Ref-COD methods is almost consistent with that of their COD counterparts, demonstrating that the distinct advantages of these COD algorithms can still be preserved when applying our design to them.
}

\myPara{\modifying{Application to Existing SOD Methods.} }
\modifying{
To further verify the generality of our Ref-COD design on Salient Object Detection (SOD) task, we also conduct experiments on the CoSOD3K~\cite{fan2021re} dataset to apply the design of our Ref-COD on the latest SOD models, \ie VSCode~\cite{luo2024vscode} and VST++~\cite{liu2024vst++}.
As shown in \tabref{tab:latest_sod_comp}, the Ref-SOD variants achieve better performance across all metrics compared to the original SOD methods.
}

\subsection{Ablative Studies} \label{subsec:abl}

\begin{figure}[t]
    \centering
    \begin{overpic}[width=0.9\linewidth]{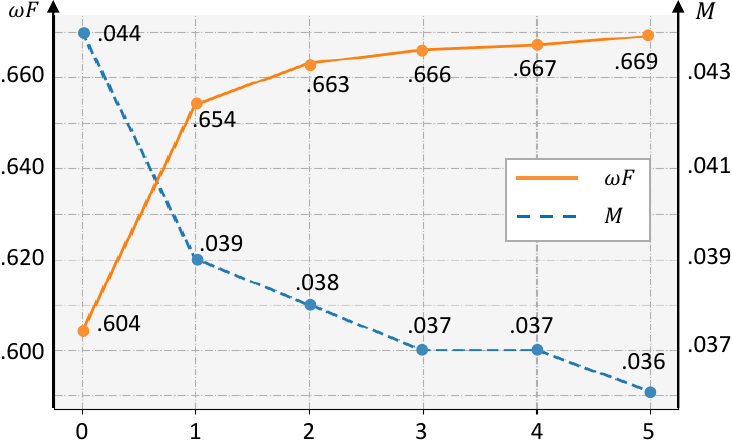}
            \put (50,0){\footnotesize{$K$}}
    \end{overpic}
    \vspace{4pt}
   \caption{\modifying{Ablation studies on the number of referring images.
   The blue dashed line and orange solid line represent the variation of two COD metrics, \ie $w$F and $M$, with the number of reference images.}}
\label{fig:abl-salnum}
\end{figure}

\myPara{Number of referring images.} 
We ablate the number of referring images, \emph{i.e.} $K$, for its impact on the \nameofbenchmark{}. Considering that there are 5 referring images in each category of the test split of the \nameofdataset{} dataset, the value of $K$ changes from 0 to 5. 
When `$0 < K < 5$', the performance of our \nameofmethod{} is calculated by averaging three evaluations, in which this model is guided by $K$ referring images randomly sampled from a certain category of the test split. 
The experimental results in \figref{fig:abl-salnum} show that the performance of our \nameofmethod{} improves with the increase of the number of referring images. We believe this is because the acquired common information is less affected by sample differences during this process, and thus a more accurate location of camouflaged targets is achieved.

\begin{table*}[tp]
  \setlength\tabcolsep{8.7pt}
  \centering
  \arrayrulecolor{\nameofmodifycolor}
  \caption{
    \modifying{
    Ablation studies on the components of our \nameofmethod{}. 
    RMG: Referring Mask Generation Moudle. 
    RFE: Referring Feature Enhance Module. 
    Both performance and computational overhead are taken into consideration.}
  }
  \label{tab:abl-var}
  \begin{tabular}{lcc|cccc|ccc} \toprule
    No. & RMG & RFE & S$_m \uparrow$ & $\alpha $E $\uparrow$ & $w$F $\uparrow$ &M$\downarrow$ & \modifying{Params (M)} & \modifying{MACs (G)} & \modifying{FPS} \\ \midrule
    1 & \xmark & \xmark & 0.772 & 0.847 & 0.604 & 0.044 & \modifying{25.97} & \modifying{21.02} & \modifying{185.74}\\
    2 & \cmark & \xmark & 0.800 & 0.870 & 0.661 & 0.038 & \modifying{26.40} & \modifying{22.30} & \modifying{169.36}\\
    3 & \xmark & \cmark & 0.792 & 0.869 & 0.644 & 0.040 & \modifying{26.55} & \modifying{21.95} & \modifying{166.57}\\
    4 & \cmark & \cmark & \textbf{0.805} & \textbf{0.879} & \textbf{0.669} & \textbf{0.036} & \modifying{27.15} & \modifying{23.23} & \modifying{151.47} \\ \bottomrule
  \end{tabular}
\end{table*}
\arrayrulecolor{black}

\begin{table}[tp]
\setlength\tabcolsep{6.7pt}
\centering
\caption{Ablation experiments on the fusion strategies in the RMG module.
`DSF': the fusion between the common representations and visual features; 
`MSF': the fusion of features at different scales.
`$\mathcal{F}_{multiply}$': element-wise multiplication. `$\mathcal{F}_{at}(\cdot)$': affine transformation.
`$\mathcal{F}_{concate}$': concatenation. `$\mathcal{F}_{clstm}(\cdot)$': convolutional lstmcell. 
 }
\begin{tabular}{lllcccc}
\toprule
No. & DSF & MSF & S$_m$ $\uparrow$ & $\alpha $E $\uparrow$  & $w$F  $\uparrow$ & M$\downarrow$\\  
\midrule 
1 &  $\mathcal{F}_{multiply}$ & $\mathcal{F}_{concate}$                         & 0.801 & 0.870 & 0.656 & 0.039 \\
2 & $\mathcal{F}_{multiply}$ & $\mathcal{F}_{clstm}$                    & 0.802 & 0.872 & 0.659 & 0.038 \\
3 & $\mathcal{F}_{at}$ &  $\mathcal{F}_{concate}$   & 0.804 & 0.872 & 0.666 & 0.037 \\
\rowcolor[HTML]{EFEFEF}
4 &  $\mathcal{F}_{at}$ & $\mathcal{F}_{clstm}$  & \textbf{0.805} & \textbf{0.879} & \textbf{0.669} & \textbf{0.036}\\
\bottomrule
\end{tabular}
\label{tab:abl-fusion}
\end{table}

\begin{table}[tp]
  \setlength\tabcolsep{5.9pt}
  \centering
  \caption{Ablation experiments on the feature enrichment strategy in the 
    RFE module. 
    `w/o CSP' or `w/ CSP': multi-scale enrichment without or 
    with the cross-scale path. 
  }
  \begin{tabular}{llccccc} \toprule
    No. & Setting & Speed (FPS) & S$_m \uparrow$ & $\alpha $E $\uparrow$ &$w$F  $\uparrow$ & M$\downarrow$ \\  \midrule
    1 & ASPP~\cite{chen2017deeplab} & 143.07 & 0.803 & 0.871 & 0.663 & 0.038 \\
    2 & RFB~\cite{liu2018receptive} & \textbf{144.03} & 0.801 & 0.872 & 0.662 & 0.038 \\
    3 & w/o CSP                    & 134.13 & 0.803 & 0.873 & 0.665 & 0.037 \\
    \rowcolor[HTML]{EFEFEF}
    4 & w/ CSP & 130.95 & \textbf{0.805} & \textbf{0.879} & \textbf{0.669} & \textbf{0.036} \\ \bottomrule
  \end{tabular}
\label{tab:abl-enrich}
\end{table}

\myPara{Model Components.} 
We analyze the importance of each component in our \nameofmethod{}, and the experimental results are shown in \tabref{tab:abl-var}. 
It can be observed that both of our modules can boost the performance of our model, where the RMG module improves the $w$F score from 0.604 to 0.661 and the RFE module improves this score from 0.604 to 0.644. 
And a more outstanding performance improvement (from 0.604 to 0.669, \emph{i.e.}, 10.8\%) appears when we combine them together.
%
%
\modifying{
These results confirm both of our proposed modules can boost the performance of our model in segmenting camouflaged objects according to the referring images with limited computational overhead.
}

\myPara{Ablation on RMG module.} 
We provide two variants for both the cross-source fusion and the cross-scale fusion of the referring representations and visual features in our RMG module. As shown in \tabref{tab:abl-fusion}, our chosen strategy achieves the best performance among the four combinations. 

\myPara{Ablation on RFE module.}
We investigate four feasible strategies for feature enrichment in our module. 
These methods include two plug-and-play fashions and two variants of our multi-scale enrichment manner.
As can be observed in \tabref{tab:abl-enrich}, our well-designed strategies achieve better performance under reasonable inference speed, especially the one with the cross-scale path.

\begin{table}[tp]
  \setlength\tabcolsep{3.5pt}
  \centering
  \caption{Ablation experiments on the channel number of our \nameofmethod{}.}
  \begin{tabular}{llcccccc} \toprule
    No. & Setting & Macs (G) & Params (M) & S$_m \uparrow$ & $\alpha $E $\uparrow$  & $w$F $\uparrow$ & M$\downarrow$ \\ \midrule
    1 & $c_{d} = 32$ & \textbf{10.6} & \textbf{24.0} & 0.799 & 0.871 & 0.655 & 0.039  \\ 
    2 & $c_{d} = 64$ & 11.7 & 25.0 & 0.805 & 0.879 & 0.669 & \textbf{0.036} \\
    3 & $c_{d} = 128$ & 15.8 & 29.1 & 0.807 & 0.875 & 0.672 & 0.037 \\
    4 & $c_{d} = 256$ & 32.2 & 44.4 & \textbf{0.811} & \textbf{0.884} & \textbf{0.679} & \textbf{0.036} \\ \bottomrule
  \end{tabular}
  \label{tab:abl-channel}
\end{table} 

\myPara{Model Dimension.} 
The dimension of the model ($c_{d}$) has a significant impact on its parameter scale and inference speed, and we thus conduct several experiments to select a suitable dimension value. 
Specifically, we increase $c_{d}$ from 32 to 256 in a doubling fashion, and the changes in model parameters, computational cost, and performance during this process are shown in \tabref{tab:abl-channel}. 
It can be observed that the performance of our \nameofmethod{} continues to improve during the increase of  $c_{d}$. However, when $c_{d}$ is greater than 64, the improvement of performance begins to decrease, while the parameters and calculation cost of the model increase sharply. 
For example, when $c_{d}$ increases from 64 to 128, the performance of our \nameofmethod{} improves slightly, but the computational overhead increases by 35\% (No.2 \emph{vs.} No.3).
Therefore, we set $c_{d}$ to 64 for the trade-off between the performance and the efficiency of our \nameofmethod{}.

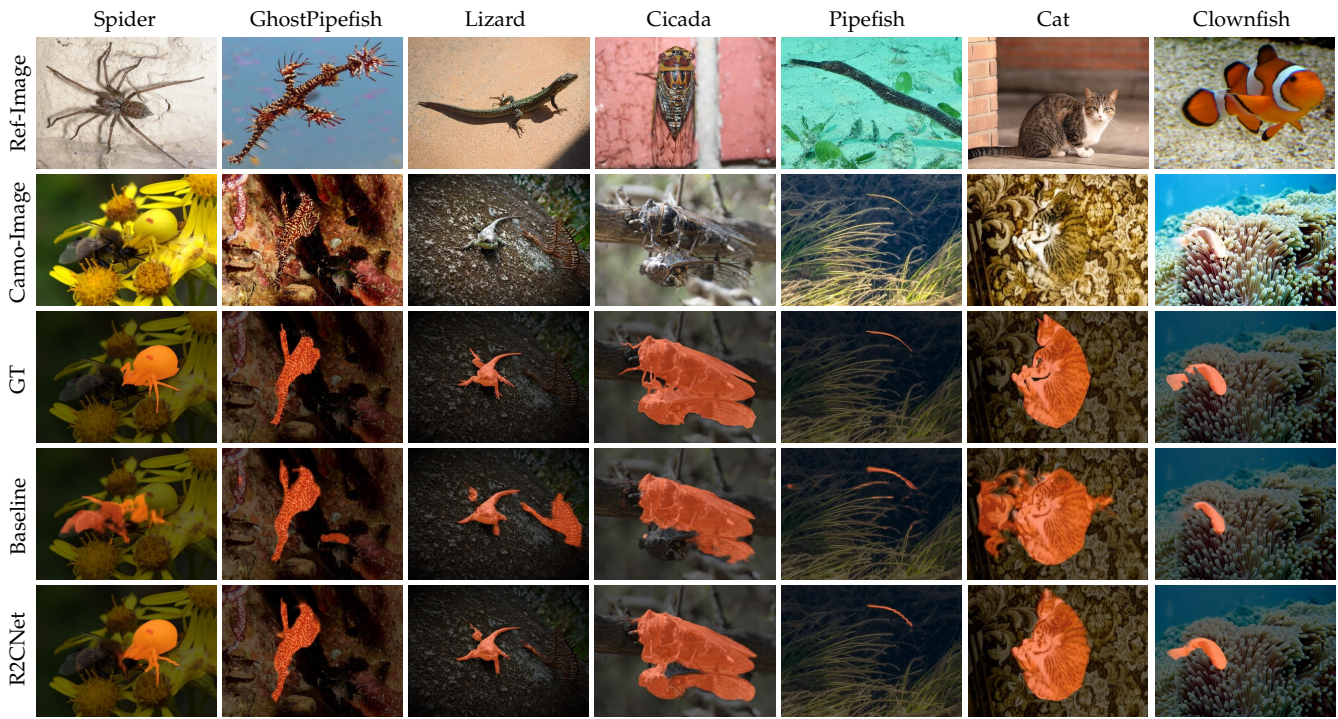
\begin{figure*}
    \centering
    \footnotesize
    \begin{overpic}[width=0.9\linewidth]{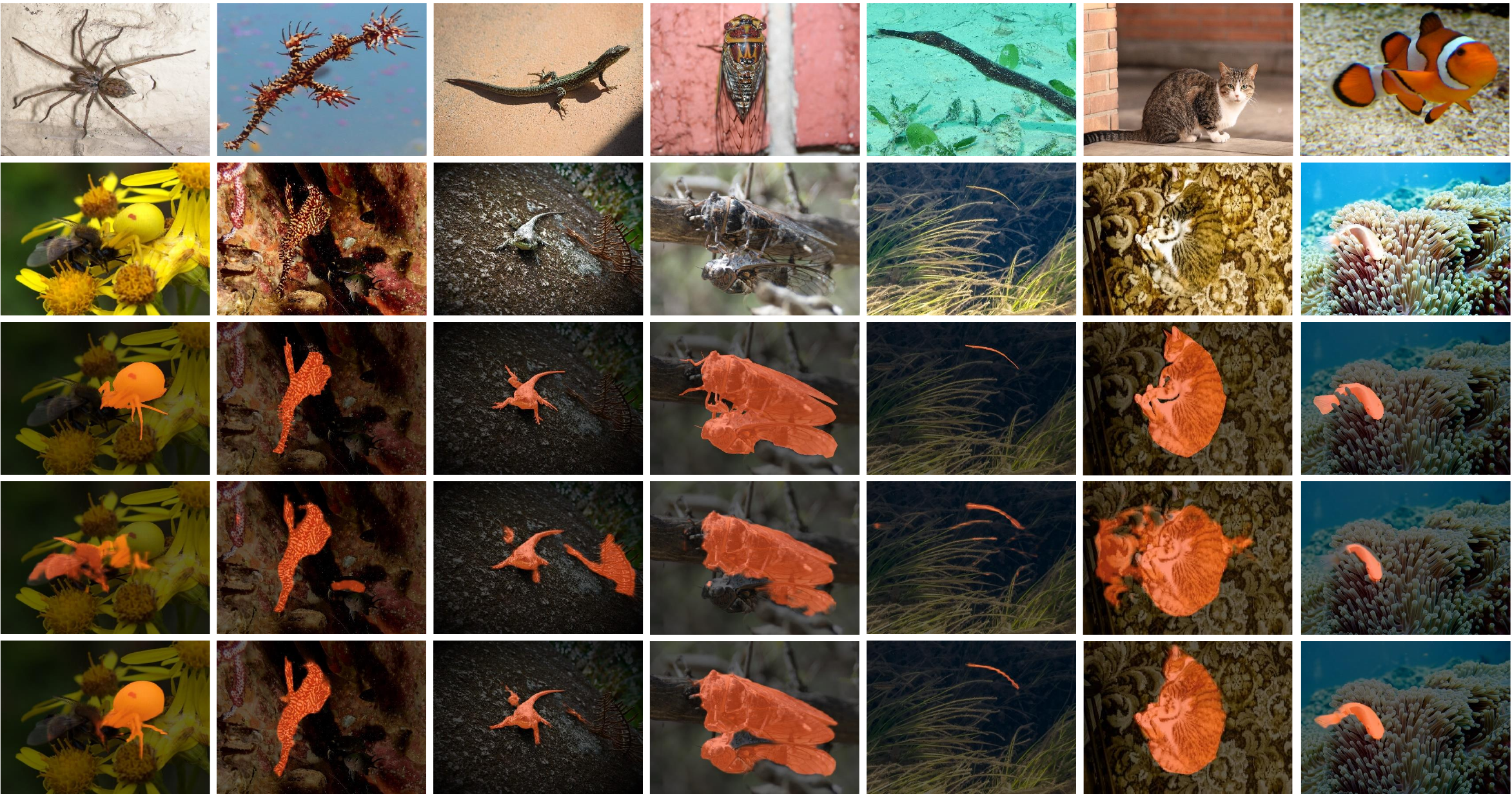}
        \put (-2,43.6){\rotatebox{90}{Ref-Image}}
        \put (-2,32){\rotatebox{90}{Camo-Image}}
        \put (-2,25){\rotatebox{90}{GT}}
        \put (-2,12.6){\rotatebox{90}{Baseline}}
        \put (-2,2.5){\rotatebox{90}{\nameofmethod{}}}
        \put (4.5,53.3){Spider}
        \put (16.5,53.3){GhostPipefish}
        \put (33,53.3){Lizard}
        \put (47,53.3){Cicada}
        \put (61,53.3){Pipefish}
        \put (77,53.3){Cat}
        \put (89,53.3){Clownfish}
    \end{overpic}
   \caption{Visual comparisons of the predictions between the proposed \nameofbenchmark{} method (\nameofmethod{}) and the standard COD method (Baseline). 
   The segmentation masks are shown in purple. 
}
\label{fig:qual-res}
\end{figure*}

\begin{table}[tp]
  \setlength\tabcolsep{4.5pt}
  \centering
  \caption{Ablation experiments on the form of \nameofbenchmark{}. text-ref: text with a simple prompt; camo-ref: image with camouflaged objects; sal-ref: image with salient objects; LSM: using large-scale pre-trained model; GT: using GT mask of the reference; Seen: the reference representations used in training appears in testing.}
  \begin{tabular}{llccccccc} \toprule
    No. & Methods & LSM & GT  & Seen & S$_m \uparrow$ & $\alpha $E $\uparrow$  & $w$F  $\uparrow$ & M$\downarrow$\\  \midrule
    1 & Baseline & & & & 0.772 & 0.847 & 0.604 & 0.044 \\
    2 & + text-ref & \cmark & & \cmark & \textbf{0.805} & 0.872 & 0.661 & 0.038 \\
    3 & + camo-ref & & \cmark & & 0.801 & 0.869 & 0.656 & 0.039 \\
    \rowcolor[HTML]{EFEFEF}
    4 & + sal-ref & & & & \textbf{0.805} & \textbf{0.879} & \textbf{0.669} & \textbf{0.036}\\
   \bottomrule
\end{tabular}

\label{tab:abl-form}
\end{table}

\myPara{Reference Forms.} 
We also investigate three different forms of reference information, \emph{i.e.}, text descriptions, images with camouflaged objects, and images with salient objects.
As discussed in \secref{sec:introduction} and \secref{sec:relatedwork}, it is difficult to obtain the detailed text descriptions and the annotated images containing the camouflaged objects associated with the camouflaged targets in the given images. 
Here we consider the readily available versions of text descriptions and images with camouflaged objects, and compare their upper bounds with our chosen reference, \emph{i.e.}, images with salient objects.

For text reference, inspired by recent prompt engineering, 
we first build text descriptions with `a photo of [CLASS]', where `[CLASS]' denotes one of the 64 categories in our \nameofdataset{}. 
Then, we feed these 64 text descriptions into the pre-trained CLIP~\cite{radford2021learning} that adopts ResNet-50 as visual backbone, to get the textual features, following CLIPSeg~\cite{luddecke2022image}.
These features are used as the common representations of target objects. 
Note that we adopt the same text representations for the samples belonging to the same category in the training set and testing set.

For image reference with camouflaged objects, we first randomly sample several images from the \camsubset{}subset of the category to which the target objects belong.
Then, we send them into ResNet-50 to extract visual features.
Note that these features are masked and pooled with ground-truth annotations to get the common representations. 

The experimental results are shown in \tabref{tab:abl-form}.
Note that the text reference employs a large-scale pre-trained model and the common representations in testing have been seen during training.
The camouflage candidate adopts GT annotations as masks to capture the common representations.
Despite these, our \nameofmethod{} using salient object reference still achieves better performance than them. 
%
%
These experiments indicate that referring images with salient objects can more effectively discover and segment the camouflaged objects.

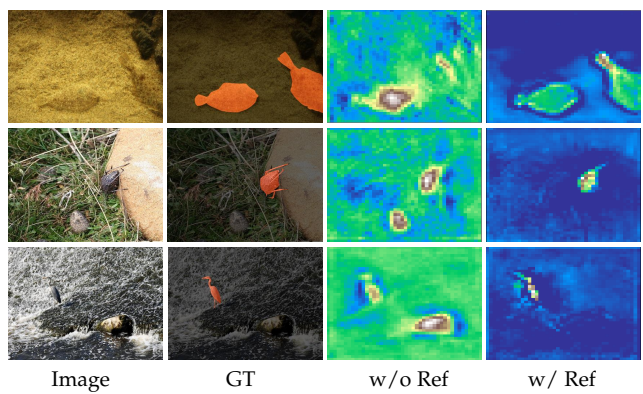
\begin{figure}[t]
    \centering
    \begin{overpic}[width=0.95\linewidth]{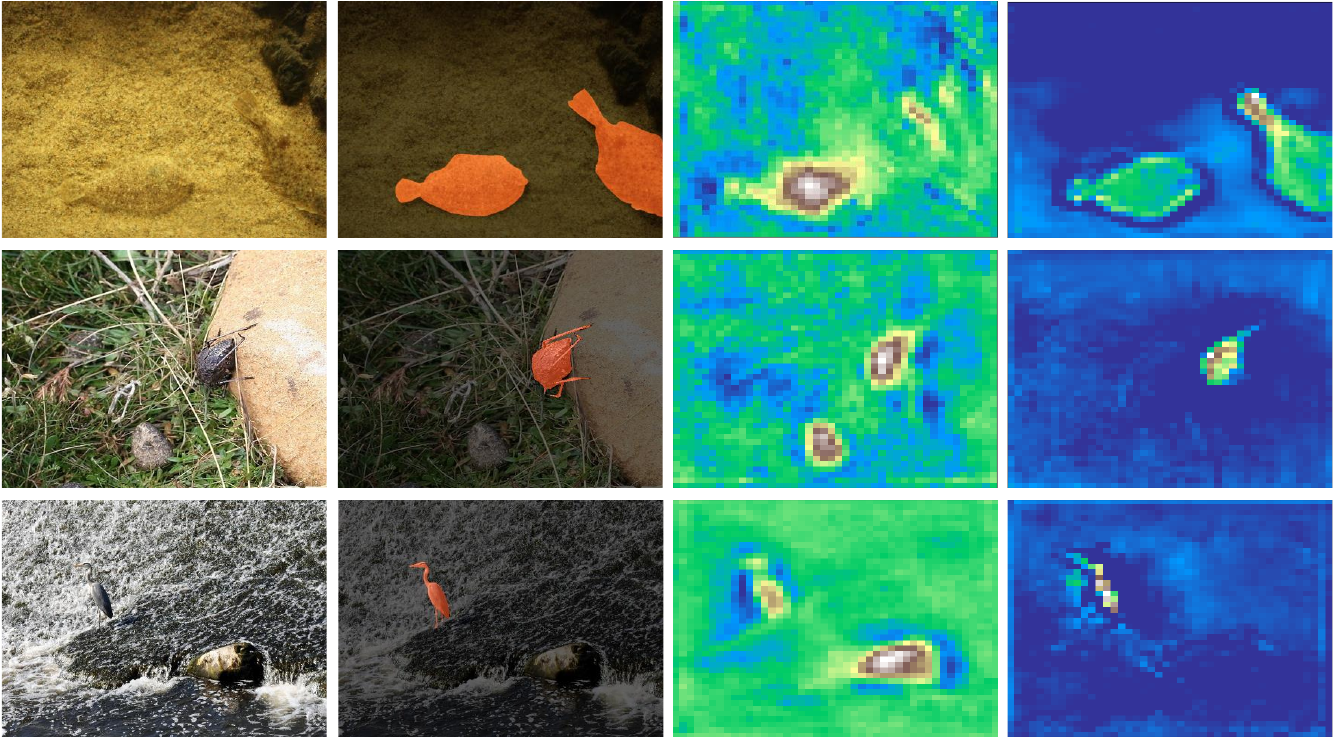}
        \put (7,-4){\footnotesize{Image}}
        \put (34.5,-4){\footnotesize{GT}}
        \put (57,-4){\footnotesize{w/o Ref}}
        \put (82,-4){\footnotesize{w/ Ref}}
    \end{overpic}
    \vspace{4pt}
   \caption{
        Visual comparison of the intermediate features between the methods with or without reference.
    }
\label{fig:qual-feat}
\end{figure}

\subsection{Qualitative Evaluation} \label{subsec:qual}

\myPara{Results Visualization.} 
We first present the visual comparison of prediction results of our referring COD method, \emph{i.e}, \nameofmethod{}, and the class-agnostic COD method, \emph{i.e}, Baseline, in \figref{fig:qual-res}. As shown in these scenes from the $1^{st}$ (Spider) to the $4^{th}$ column (Cicada), there are multiple camouflaged objects. 
The baseline model tends to indiscriminately segment all potential camouflaged objects, and thus its outputs are often inaccurate.
To be specific, it may not only miss some camouflaged objects (\emph{e.g.}, Cicada), but also mispredict other objects as camouflaged ones (\emph{e.g.}, Spider and Lizard).
For all these challenging scenes, our \nameofmethod{} is able to identify the camouflaged objects orienting to referring images. 
In addition, compared with the baseline model, our \nameofmethod{} is less disturbed by other similar objects (\emph{e.g.}, Pipefish). A nd it can segment visual objects more completely in scenes with only one camouflaged object but with less difference from the surrounding, as shown in the $6^{th}$ column (Cat) and $7^{th}$ column (Clownfish). 
We argue that this is because the common representations of a certain category of objects learned from the referring images are conducive to identifying the main body of such camouflaged objects.

\myPara{Features Visualization.} 
To gain more insights from the \nameofbenchmark{}, we also visualize intermediate features with or without referring information, as shown in \figref{fig:qual-feat}.
As can be seen, the model without references can locate the camouflaged objects but it may be disturbed by other objects, and produces inaccurate segmentation results. With the incorporation of references, the model can focuse more on the indicated camouflaged objects, regardless of other ones.

\section{Future Work}
Based on the \nameofdataset{} dataset and \nameofmethod{} framework proposed in this paper, there are a few foreseeable directions about the \nameofbenchmark{} to explore in future research:


\textbf{1) Other Forms of References.} 
In recent years, multi-modal research has made great progress in vision and language, vision and speech, \emph{etc}. Based on them, it is interesting to explore \nameofbenchmark{} with other references, \emph{e.g.}, text and speech. 
As discussed above, the main challenge depends on how to obtain these references easily and efficiently. 

\textbf{2) Special Case of Referring Scene.} 
In this paper, we assume the image to be segmented contains the target objects specified by the referring images. 
However, such objects may not exist in some real scenes.
It will be promising to extend \nameofbenchmark{} to be compatible with these cases.

\textbf{3) Related Tasks.} 
In some application scenarios, users only need some simple answers to make decisions, rather than using more complex equipment to obtain the segmentation details of camouflaged objects. Therefore, it is also valuable to extend the \nameofbenchmark{} to simple tasks related to question answering.

\section{Conclusions}
We propose a novel benchmark named \nameofbenchmark{} to directionally segment the camouflaged object with simple image references. 
First, we build a large-scale dataset (\emph{i.e.}, \nameofdataset{}) consisting of images in real-world scenarios. 
Then, we develop an end-to-end Reference to COD framework (\emph{i.e.}, \nameofmethod{}), which follows the two-branch architecture. 
Particularly, the introduction of our well-designed RMG module and RFE module facilitates the guidance of the common representations from referring images to the challenging segmentation of camouflaged objects. 
Compared with the COD counterpart which indiscriminately segments camouflaged objects from the background, our \nameofmethod{} achieves significantly superior performance on the common metrics, and generates more visually favorable predictions.
We believe it is promising to establish a multi-information collaborative system, and we hope the perspective provided by our \nameofbenchmark{} will inspire future work.

\bibliography{main}

\bibliographystyle{IEEEtran}

\newcommand{\AddPhoto}[1]{{\includegraphics[width=1in,keepaspectratio]{authors/#1}}}
\newcommand{\AuthorBio}[3]{\vspace{-.25in}\begin{IEEEbiography}[\AddPhoto{#1}]{#2}#3\end{IEEEbiography}}

\AuthorBio{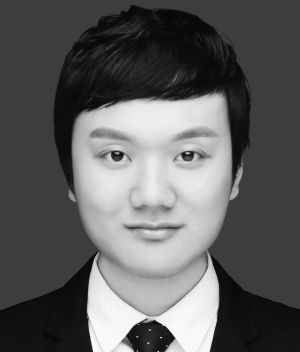}{Xuying Zhang}{
is a Ph.D. student at the College of Computer Science, Nankai University, 
under the supervision of Prof. Ming-Ming Cheng.
His research interests include multimodal learning, camouflaged scene understanding, and 2D/3D content generation.
}

\AuthorBio{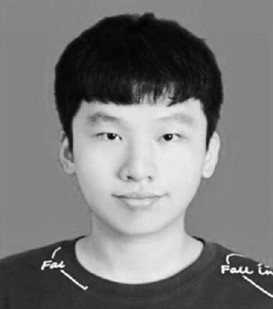}{Bowen Yin}{is a Ph.D. student from the college of computer science, Nankai university.
He is supervised by Prof. Qibin Hou.
His research interests include computer vision and multimodal scene perception.
}

\AuthorBio{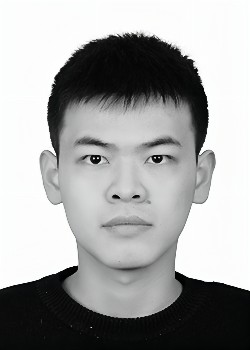}{Zheng Lin}{
is currently a postdoctoral researcher at Tsinghua University, under the supervision of Prof. Shi-Min Hu. He get the doctoral degree from Nankai University in 2023, under the supervision of Prof. Ming-Ming Cheng. His research interest is in computer vision and computer graphics.
}

\AuthorBio{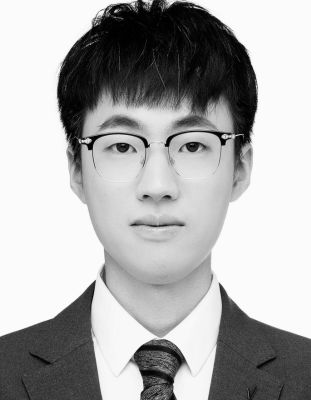}{Qibin Hou}{(Member, IEEE) 
received his Ph.D. degree from the School of Computer Science, 
Nankai University. Then, he spent two wonderful years working at the National University of Singapore as a research fellow. 
Now, he is an associate professor at School of Computer Science, Nankai University. 
He has published more than 40 papers on top conferences/journals, including IEEE TPAMI, CVPR, ICCV, NeurIPS, etc. 
His research interests include deep learning and computer vision.  }

\AuthorBio{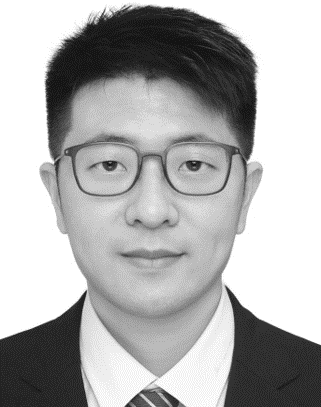}{Deng-Ping Fan}{(Senior Member, IEEE)
will be joining the Department of Nankai International Advanced Research Institute (SHENZHEN-FUTIAN) as a faculty member in 2024. Now I'm a Full Professor and deputy director of the Media Computing Lab (MC Lab) at the College of Computer Science, Nankai University, China. Before that, he was postdoctoral, working with Prof. Luc Van Gool in Computer Vision Lab @ ETH Zurich. He is one of the core technique members in TRACE-Zurich project on automated driving. 
}

\AuthorBio{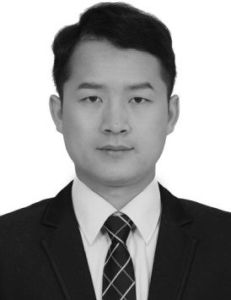}{Ming-Ming Cheng}{(Senior Member, IEEE) received his PhD degree from Tsinghua University in 2012,
and then worked with Prof. Philip Torr in Oxford for 2 years.
Since 2016, he is a full professor at Nankai University, leading the Media Computing Lab. 
His research interests include computer vision and computer graphics.
He received awards, including ACM China Rising Star Award,
IBM Global SUR Award, \etc.
He is a senior member of the IEEE and on the editorial boards of
IEEE TPAMI and IEEE TIP.
}

\vfill

\end{document}